\documentclass{article}
\usepackage{microtype}
\usepackage{graphicx}
\usepackage{subcaption}
\usepackage{booktabs}
\usepackage{hyperref}

\usepackage[preprint]{icml2026}
\usepackage{amsmath}
\usepackage{amssymb}
\usepackage{mathtools}
\usepackage{amsthm}
\usepackage{multirow}
\usepackage{capt-of}
\usepackage{algorithm}
\usepackage{algorithmic}
\usepackage{colortbl}
\usepackage{tikz}
\usetikzlibrary{patterns,positioning,calc,arrows.meta}
\usepackage{pgfplots}
\pgfplotsset{compat=1.17}
\usepgfplotslibrary{groupplots}
\usepackage[capitalize,noabbrev]{cleveref}
\theoremstyle{plain}
\newtheorem{proposition}{Proposition}
\theoremstyle{definition}

\definecolor{cPL}{HTML}{2A78D6}
\definecolor{cWS}{HTML}{EB6834}
\definecolor{cBQ}{HTML}{1BAF7A}
\definecolor{cDone}{HTML}{C9C8C1}
\definecolor{cInk}{HTML}{3D3D3A}

\newcommand{\method}{\textsc{PackLoop}}
\newcommand{\ours}{\method{} (ours)}

\newcommand{\oursrow}{\rowcolor{cPL!9}}
\newcommand{\codenote}{}

\icmltitlerunning{PackLoop: Row-Packed Self-Speculative Decoding for Looped Language Models}

\begin{document}

\twocolumn[
  \icmltitle{PackLoop: Row-Packed Self-Speculative Decoding\\for Looped Language Models under Concurrency}
  \icmlsetsymbol{equal}{*}
  \begin{icmlauthorlist}
    \icmlauthor{Runguo Li}{uiuc}
    \icmlauthor{Dezhen Lu}{tue}
  \end{icmlauthorlist}
  \icmlaffiliation{uiuc}{University of Illinois Urbana-Champaign}
  \icmlaffiliation{tue}{University of T\"ubingen}
  \icmlcorrespondingauthor{Runguo Li}{\mbox{runguol2@illinois.edu}}
  \icmlkeywords{Looped language models, speculative decoding, inference, serving}
  \vskip 0.2in
  \begin{center}
  \input{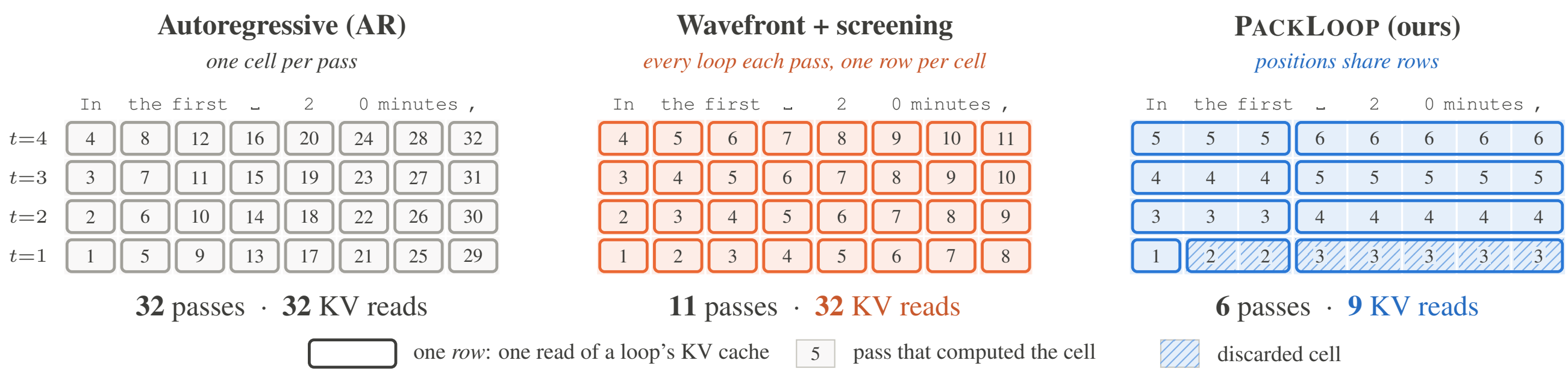}
  \vskip 0.08in
  \captionof{figure}{\textbf{Speculate only inside rows that are read anyway.} The first eight tokens of a GSM8K answer from Ouro-1.4B ($T{=}4$ loops) under three schedules, replayed from recorded per-loop readouts. Each outlined box is a \emph{row}, one read of that loop's KV cache, which is what a decoding pass costs under concurrency. The screened wavefront advances every loop in each pass, yet every cell still occupies a row of its own. \method{} lets positions share rows: a proposal copied from the question joins the row that loop~1 reads in any case, and the deep loops advance in blocks. Rejected proposal tails cost cells but no row.}
  \label{fig:grid}
  \end{center}
  \vskip 0.1in
]
\printAffiliationsAndNotice{}

\begin{abstract}
Looped language models reuse one block of weights $T$ times per token, trading parameters for test-time depth, but autoregressive decoding then pays $T$ sequential passes per token. Self-speculative decoders recover 2--10$\times$ for a single request, yet their gains fade as requests share a GPU, and current practice is to speculate less as load grows. We show that what fails under load is the \emph{layout} of speculative work, not speculation. In concurrent decoding the unit of cost is the \emph{row}, one request advancing on one loop, which reads that loop's key--value (KV) cache once; additional positions inside a row are nearly free. Pipelined schedules such as the wavefront place every speculative position on a different loop, hence in a row of its own, so each token is still read $T$ times. We present \method{}, built on one principle: speculate only inside rows that are read anyway. Deep loops accumulate positions and advance them as a block in one row, context-copy proposals join the row that the first loop already reads, and a closed-form admission rule, the economic order quantity of inventory theory, sets how long to accumulate from a cost model calibrated once per engine. \method{} is exact: in FP32 its output is token-for-token identical to autoregressive decoding, also under sampling and early exit. Across 164 load points spanning three GPUs, four Ouro models, six benchmarks, and contexts up to 32K tokens, one \method{} configuration is faster than the best existing schedule, chosen per point, at 161 points (median 1.25$\times$, up to 1.76$\times$) and never significantly slower. Its lead grows with concurrency and context length, the regime in which existing self-speculation gains least.\codenote
\end{abstract}

\section{Introduction}
\label{sec:intro}

Looped (recurrent-depth) language models apply one shared block of layers $T$ times to every token \citep{universal,huginn,ouro}. They trade parameters for test-time depth, and Ouro matches considerably larger non-looped models \citep{ouro}. The price is paid at decoding time: every token needs $T$ sequential passes through the block, each with its own key--value (KV) cache, so serving cost decides whether such models are practical.

For a single request, self-speculative decoding removes most of this cost. A shallow loop's readout drafts the next token, deeper loops verify it, and positions at different depths share one forward pass. WaveFront \citep{wavefront}, LoopSpec \citep{loopspec}, LoopAhead \citep{loopahead}, CopyLoop \citep{copyloop}, and the concurrent LoopFill \citep{loopfill} and LoopSkim \citep{loopskim} report speedups from 2.3$\times$ to more than 10$\times$ with one request per GPU. Their cost models count sequential passes.

\begin{figure*}[t]
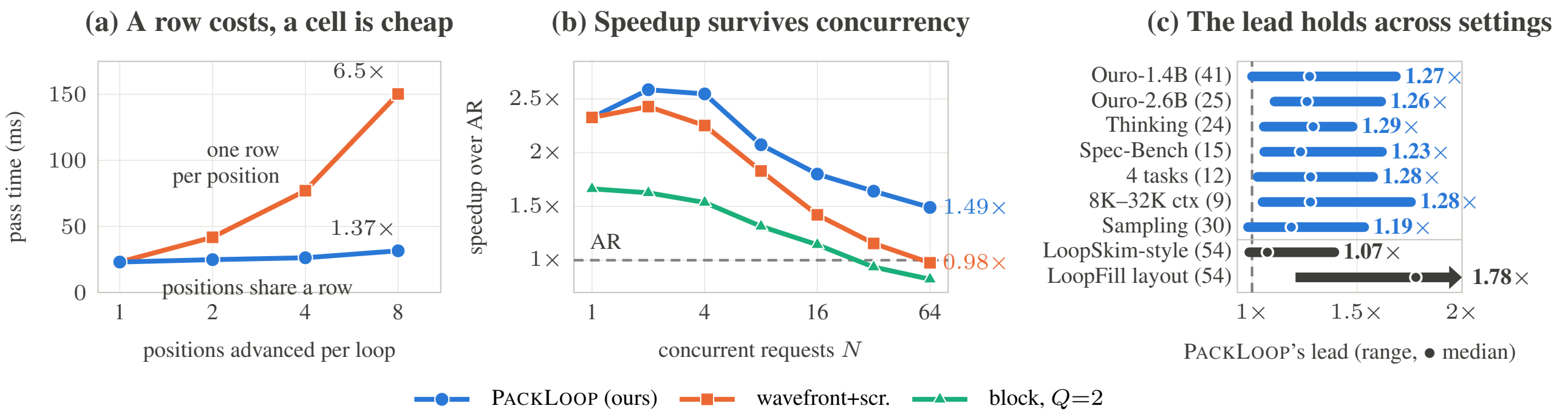

\centering
\input{fig_teaser_data.tex}
\begin{tikzpicture}
\pgfplotsset{
  teaser/.style={
    height=4.1cm,
    tick label style={font=\scriptsize, color=cInk},
    label style={font=\scriptsize, color=cInk},
    title style={font=\small\bfseries, color=cInk, align=center, yshift=-2pt},
    grid=major, grid style={color=cDone!45, line width=0.3pt},
    axis line style={color=cDone}, major tick length=0pt, minor tick length=0pt, xtick pos=bottom, ytick pos=left, clip=false,
    every axis plot/.append style={line width=1.5pt},
  },
}
\begin{axis}[teaser, name=pa, width=0.31\textwidth,
  title={(a) A row costs, a cell is cheap},
  xmode=log, log basis x=2, xtick={1,2,4,8}, xticklabels={1,2,4,8}, xmin=0.85, xmax=11,
  ymin=0, ymax=175, ytick={0,50,100,150},
  xlabel={positions advanced per loop}, ylabel={pass time (ms)},
]
\addplot[color=cWS, mark=square*, mark size=2pt, mark options={draw=white, line width=0.4pt}] coordinates {\TAfl};
\addplot[color=cPL, mark=*, mark size=2.2pt, mark options={draw=white, line width=0.4pt}] coordinates {\TApk};
\node[font=\scriptsize, text=cInk, anchor=east, align=right] at (axis cs:3.6,95) {one row\\per position};
\node[font=\scriptsize, text=cInk, anchor=north] at (axis cs:2.8,19) {positions share a row};
\node[font=\scriptsize\bfseries, text=cInk, anchor=south east] at (axis cs:8,152) {$6.5\times$};
\node[font=\scriptsize\bfseries, text=cInk, anchor=south east] at (axis cs:8.6,33) {$1.37\times$};
\end{axis}
\begin{axis}[teaser, name=pb, at={($(pa.east)+(1.45cm,0)$)}, anchor=west, width=0.33\textwidth,
  title={(b) Speedup survives concurrency},
  xmode=log, log basis x=2, xtick={1,4,16,64}, xticklabels={1,4,16,64}, xmin=0.8, xmax=80,
  legend style={font=\scriptsize, draw=none, fill=none, at={(0.5,-0.36)}, anchor=north, legend columns=3, column sep=4pt},
  ymin=0.7, ymax=2.85, ytick={1,1.5,2,2.5}, yticklabels={$1\times$,$1.5\times$,$2\times$,$2.5\times$},
  xlabel={concurrent requests $N$}, ylabel={speedup over AR},
]
\addplot[gray, densely dashed, line width=0.8pt, mark=none, forget plot] coordinates {(0.8,1) (80,1)};

\addplot[color=cPL, mark=*, mark size=2.2pt, mark options={draw=white, line width=0.4pt}] coordinates {\TBpl};
\addlegendentry{\method{} (ours)}
\addplot[color=cWS, mark=square*, mark size=2pt, mark options={draw=white, line width=0.4pt}] coordinates {\TBws};
\addlegendentry{wavefront+scr.}
\addplot[color=cBQ, mark=triangle*, mark size=2.4pt, mark options={draw=white, line width=0.4pt}] coordinates {\TBbq};
\addlegendentry{block, $Q{=}2$}
\node[font=\scriptsize, text=cInk, anchor=south west] at (axis cs:0.85,1.0) {AR};
\node[font=\scriptsize\bfseries, text=cPL, anchor=west, inner sep=1.5pt] at (axis cs:70,1.491) {$1.49\times$};
\node[font=\scriptsize\bfseries, text=cWS, anchor=west, inner sep=1.5pt] at (axis cs:70,0.976) {$0.98\times$};
\end{axis}
\input{fig_teaser_c_data.tex}
\begin{axis}[teaser, name=pc, at={($(pb.east)+(3.15cm,0)$)}, anchor=west, width=0.235\textwidth,
  title={(c) The lead holds across settings},
  xmin=0.93, xmax=2.0, xtick={1,1.5,2}, xticklabels={$1\times$,$1.5\times$,$2\times$},
  ymin=0.4, ymax=9.6, ytick=\empty, ymajorgrids=false,
  xlabel={\method{}'s lead (range, $\bullet$ median)},
]
\draw[gray, densely dashed, line width=0.8pt] (axis cs:1,0.4) -- (axis cs:1,9.6);
\draw[cDone, line width=0.4pt] (axis cs:0.93,2.5) -- (axis cs:2.0,2.5);
\TDbars
\end{axis}
\end{tikzpicture}
\caption{The argument in three measurements (Ouro-1.4B). (a) \emph{Cause.} One decoding pass in which all four loops advance, 4 requests at context 4096 on an RTX 3090: eight positions in each loop's row take 1.37$\times$ the time of one, eight single-position rows take 6.5$\times$. (b) \emph{Consequence.} Speedup over AR on the H100 at context 1024: the screened wavefront and blocks decay to AR speed at 64 requests, \method{} retains 1.49$\times$. (c) \emph{Breadth.} \method{}'s lead over the per-point fastest existing schedule in the same engine, by setting (load points in parentheses); last two rows: over re-implementations of LoopSkim and LoopFill (\cref{sec:external}).}
\label{fig:teaser}
\end{figure*}

Serving is concurrent, and under concurrency these gains fade. LoopFill's speedup falls from 4.9$\times$ with one request to 2.8$\times$ with 16 \citep{loopfill}. On the vllm-rlt serving engine, fixed-depth self-speculation turns from a 1.52$\times$ speedup with one request into a 0.75$\times$ slowdown with 32 \citep{exactloop}. In our engine, the screened wavefront falls from 2.3$\times$ to the speed of autoregressive (AR) decoding at 64 requests (\cref{fig:teaser}b). The prevailing response is to speculate less as load grows, with a draft length tuned per batch size \citep{exactloop,loopskim,vllmrlt}.

We show that this response addresses the wrong cause: what fails under load is the \emph{placement} of speculative work, not speculation. Once several requests share a GPU, the time of a decoding pass is governed by its \emph{rows}. A row is one request advancing on one loop. It reads that loop's KV cache once, at a rate close to memory bandwidth, however many positions it advances. Additional positions inside a row are therefore nearly free, whereas every additional row costs a full read (\cref{fig:teaser}a). Pipelined self-speculation cannot exploit this asymmetry. A wavefront advances position $i$ on loop $T$, position $i{+}1$ on loop $T{-}1$, and so on, so each speculative position sits on a different loop and occupies a row of its own (\cref{fig:grid}). Passes per token fall, but each token's cache is still read $T$ times, as in AR decoding, plus the reads spent on rejected drafts. A controlled experiment confirms that layout alone accounts for the collapse: with the schedule held fixed, splitting each multi-position row into one row per position turns a 1.50$\times$ speedup at 64 requests into a 0.49$\times$ slowdown (\cref{tab:reads}).

The diagnosis prescribes the remedy: \emph{speculate only inside rows that are read anyway}. Positions can share a row only when they are ready on the \emph{same} loop in the same pass, which a pipeline offers only if its deep loops wait and accumulate. \method{} (\cref{sec:method}) turns this principle into a decoder with three components:
\begin{itemize}
\item \textbf{Row-packed blocks.} Each deep loop accumulates ready positions and advances them as one block in one row. Blocks move down the loops in a staggered pipeline, so a token's cache is read about $1+(T-1)/Q$ times instead of $T$.
\item \textbf{In-row proposals.} Multi-token proposals copied from the context are placed in the row that loop~1 reads in any case, so a pass can commit several tokens per request without opening a new row.
\item \textbf{EOQ admission.} Waiting saves rows but delays the detection of rejected drafts. Under the measured cost model this trade-off is an inventory problem whose solution is the economic order quantity, $B^\star=\sqrt{2a\rho/\kappa}$. Its inputs are calibrated once per engine or estimated online, so one configuration behaves as a pipeline at light load and forms large blocks at heavy load.
\end{itemize}
\method{} is exact. Every committed token is the last loop's readout on exactly computed inputs, so FP32 outputs equal those of AR decoding token for token, including under sampling and adaptive early exit.

We evaluate \method{} on Ouro-1.4B, Ouro-2.6B, and their reasoning-tuned versions \citep{ouro} on three GPUs, with 1--192 concurrent requests, contexts of 256 to 32K tokens, six benchmarks, and greedy and sampled decoding (\cref{sec:exp}). Every existing schedule runs in the same engine and job as \method{} and is credited with its best configuration at each load point, whereas \method{} uses one configuration throughout. Across the resulting 164 load points, \method{} is faster at 161, by 1.25$\times$ at the median and up to 1.76$\times$, and it is never significantly slower (\cref{fig:teaser}c). The lead grows with concurrency and context length. \method{} also outperforms re-implementations of the concurrent LoopSkim and LoopFill, CopyLoop and LoopSpec on their own single-request protocol, and continuous depth batching \citep{cdb} under the same early-exit rule (\cref{tab:external}).

In summary, our contributions are threefold:
\begin{itemize}
\item \textbf{Analysis.} We identify rows of KV reads as the unit of cost in concurrent looped-model decoding and show that pipelined self-speculation is structurally unable to share rows, which explains why its gains vanish under load (\cref{sec:cost}).
\item \textbf{Method.} We propose \method{}, an exact decoder that packs speculative positions into rows that are read anyway and whose closed-form admission rule lets a single configuration serve light and heavy load (\cref{sec:method}).
\item \textbf{Evidence.} We compare \method{} with every existing schedule in one engine and with external systems on their own protocols, and validate a replay model that extends the comparison to the 32-iteration Huginn (\cref{sec:exp}).
\end{itemize}

\section{Background}
\label{sec:background}

\textbf{Looped models and the decoding grid.} A looped model applies a block $f$ of $L$ layers $T$ times: $h^{(t)}_i = f(h^{(t-1)}_i;\, \mathrm{KV}^{(t)}_{<i})$ for $t=1,\dots,T$, where $\mathrm{KV}^{(t)}_{<i}$ are the keys and values that loop $t$ produced for earlier positions. A readout head on $h^{(t)}_i$ predicts token $i{+}1$, and the token is committed from loop $T$. Decoding fills a grid of positions $\times$ loops (\cref{fig:grid}). A \emph{cell} $(i,t)$ is the computation of position $i$ on loop $t$; it is ready once $(i,t{-}1)$ and all $(j,t)$ with $j<i$ are done. A \emph{row} is one request advancing on one loop in a pass: it reads that loop's KV cache for the request once, however many cells it advances. AR decoding fills one cell per pass. Throughout, $N$ denotes the number of concurrent requests and $P$ the context length in tokens.

\textbf{Self-speculation.} Because $h^{(1)}_i$ already gives a good guess of token $i{+}1$, position $i{+}1$ can start on loop 1 before position $i$ finishes on loop $T$; if the final readout disagrees, the cells that depended on the wrong token are discarded. A \emph{wavefront} advances every loop by one cell per pass along the anti-diagonal \citep{wavefront,loopspec}. LoopSpec adds a second proposal at a deeper loop, LoopAhead trains a drafter that proposes several future tokens \citep{loopahead}, and CopyLoop copies multi-token suffixes from the context and screens them at a shallow loop \citep{copyloop}. The \emph{block} mode of vllm-rlt drafts $K$ tokens and verifies them in one pass, with each verification token in its own row \citep{vllmrlt}. Adaptive early exit lets tokens leave before loop $T$, and continuous depth batching regroups tokens between loops so that each exits at its own depth \citep{cdb}.

\section{What Concurrent Decoding Costs}
\label{sec:cost}

Three findings form the empirical basis of \method{}: a pass is priced by its rows, existing self-speculation saves passes but not rows, and layout alone decides whether speculation pays under load.

\textbf{Finding 1: A pass is priced by its rows.} We time decoding passes in our engine while varying how many loops advance and how many cells are computed. The time of a pass is well described by
\begin{equation}
\tau \;\approx\; c_0 \;+\; \beta\cdot P_{\text{act}} \;+\; \gamma\cdot g(C),
\label{eq:cost}
\end{equation}
where $P_{\text{act}}$ is the number of loops that advance in the pass, $C$ is the number of computed cells, and $g$ is a calibrated, step-shaped function for the cost of the matrix multiplications, with scale $\gamma$. Each advancing loop contributes one row per request, so $\beta$ is the cost of reading one loop's cache for all $N$ requests, and it grows with $N$ and the context length. Fitted online, \cref{eq:cost} predicts the next pass time within 0.4--1.6\% on the H100 and 0.9--6.2\% on the RTX 3090. The row term dominates as soon as several requests share the GPU, because a cache read runs close to memory bandwidth and barely depends on how many positions consume it. With four requests at context 4096, advancing eight positions in each loop's row takes 1.37$\times$ the time of advancing one, whereas advancing them in eight single-position rows takes 6.5$\times$ (\cref{fig:teaser}a).

\begin{table}[t]
\caption{Passes and KV-cache reads per committed token on an H100 (Ouro-1.4B, $T=4$, context 1024, 64 requests), and speedup over AR with $N$ concurrent requests. The last two rows run the same \method{} schedule and differ only in layout; with split rows the admission rule pipelines, hence the fewer passes.}
\label{tab:reads}
\centering\small
\setlength{\tabcolsep}{4pt}
\resizebox{\linewidth}{!}{%
\begin{tabular}{lccccc}
\toprule
 & & & \multicolumn{3}{c}{$\times$AR at $N=$} \\
\cmidrule(l){4-6}
Schedule & passes & reads & 16 & 32 & 64 \\
\midrule
Autoregressive & 4.00 & 4.0 & 1.00 & 1.00 & 1.00 \\
Wavefront + screening & 1.31 & 5.2 & 1.44 & 1.16 & 0.98 \\
\method{}, one row per position & 0.94 & 6.6 & 0.83 & 0.60 & 0.49 \\
\oursrow \ours{} & 1.15 & \textbf{2.0} & \textbf{1.82} & \textbf{1.65} & \textbf{1.50} \\
\bottomrule
\end{tabular}}
\end{table}

\textbf{Finding 2: Pipelined speculation saves passes, not reads.} \Cref{tab:reads} counts passes and cache reads per committed token. Single-request cost models count passes, and speculation indeed cuts them from 4 to about 1. Reads behave differently. The screened wavefront still reads each token's cache once on every loop and pays further reads for rejected drafts, 5.2 reads per token against 4.0 for AR. Its speedup accordingly decays from 1.44$\times$ at 16 requests to 0.98$\times$ at 64. Only a schedule that advances several positions per row reduces reads: a block of $Q$ positions in one row brings them to about $1+(T-1)/Q$.

\textbf{Finding 3: Layout alone decides the speedup.} The last two rows of \cref{tab:reads} isolate layout. Both run \method{}'s schedule and admission rule and differ only in whether the positions that a loop advances share one row. Splitting the rows more than triples the reads per token and turns the 1.82$\times$ and 1.50$\times$ speedups at 16 and 64 requests into 0.83$\times$ and 0.49$\times$, slower than AR. This reproduces, inside one engine, the slowdown reported for per-token-row verification \citep{exactloop,vllmrlt}. The same holds on the RTX 3090, where the split leaves 18--66\% of \method{}'s throughput (\cref{tab:ablation}), and on an A6000 and an A100 (\cref{app:tables}).

\textbf{Implication: sharing rows requires waiting.} A pipeline gains speed by overlapping loops, so at any moment the positions it can advance sit on different loops, each with its own cache; a wavefront has exactly one ready cell per loop. Positions share a row only when deep loops wait until several of them are ready on the same loop, and waiting is not free, because a deferred cell also defers the detection of a wrong draft. Round-based verification, as in vllm-rlt's block mode and in non-looped speculative decoding \citep{leviathan2023,draftverify}, shares rows by construction but gives up the pipeline. The design question is therefore how to wait \emph{inside} a pipeline and for how long, and the answer depends on load, context length, and hardware.

\section{PackLoop}
\label{sec:method}

\method{} schedules cells on the decoding grid so that speculative width only enlarges rows that are read anyway. The findings of \cref{sec:cost} raise three design questions, which we answer in turn: how deep loops share rows without stalling the pipeline (\cref{sec:blocks}), how drafts longer than one token enter without a row of their own (\cref{sec:proposals}), and how long a loop should wait (\cref{sec:eoq}). \Cref{sec:exact} shows that the resulting decoder is exact, and \cref{alg:packloop} summarizes one pass.

\begin{algorithm}[t]
\caption{One \method{} pass over the batch}
\label{alg:packloop}
\begin{algorithmic}[1]
\STATE Each request: loop 1 takes its next draft position and up to $m$ copied proposal positions, in one row
\STATE Look up the deep loops' attention cost $F$ at the current context length (calibrated once per engine)
\STATE \textbf{if} $F/\bar\tau<f^\star$ \textbf{then} \emph{pipeline}: every deep loop with a backlog advances, at loop 1's width
\STATE \textbf{else} \emph{blocks}: the first deep loop $t_0$ advances if its mean backlog per request is at least $B^\star_{t_0}$ (\cref{eq:eoq-lead}); deeper loops advance whenever they have a backlog, up to $Q$ cells
\STATE Execute: linear layers on the selected cells only; one attention call per group of loops with equal width
\STATE Read out every advanced loop; where a readout disagrees with the next token, replace the token and discard the cells that depended on it
\STATE Commit tokens whose loop-$T$ readout is final
\STATE Update the online estimates of $\bar\tau$, $a$, and $\kappa$
\end{algorithmic}
\end{algorithm}

\subsection{Row-packed blocks}
\label{sec:blocks}

Loop 1 advances in every pass and drafts one token from its own readout. Each deeper loop $t>1$ accumulates a backlog of ready cells and, once admitted (\cref{sec:eoq}), advances up to $Q=8$ of them in one row. A block advanced by loop $t$ becomes ready for loop $t{+}1$ in the next pass, so blocks travel down the loops staggered by one pass and the deep loops form a pipeline of $T-1$ stages over positions. Drafting never waits for verification. When a deeper readout at position $i$ disagrees with the drafted token $i{+}1$, the draft is replaced (\emph{screening}) and the cells that depended on it are discarded. With blocks of $Q$ positions, a committed token is read once on loop 1 and about $1/Q$ times on each deep loop, which gives the $1+(T-1)/Q$ reads of \cref{sec:cost}; \cref{tab:reads} measures 2.0 at $T=4$.

\subsection{In-row proposals}
\label{sec:proposals}

With one drafted token per pass, a request commits at most one token per pass. \method{} lifts this bound with multi-token proposals copied from the context. As in CopyLoop \citep{copyloop}, it locates the most recent earlier occurrence of the current suffix (1--4 tokens) and proposes the up to $m=8$ tokens that followed it. The difference lies in where the proposal is placed. CopyLoop runs each candidate as a pipeline lane of its own, that is, in a row of its own. \method{} places the proposed positions in the loop-1 row that the request reads in any case. Loop 1's readouts screen the proposal within that row, and the accepted prefix flows into the deep-loop blocks, so a rejected tail costs cells but never a row (\cref{fig:grid}). On HumanEval with one request, \method{} needs 0.43 passes per token.

\textbf{Load-set width.} Extra cells in a row stay cheap only while the matrix multiplications remain narrow. When at least $w_n$ requests are generating, \method{} therefore caps the loop-1 row at 3 cells instead of 8. $w_n$ is the only hardware-specific constant of \method{} and is set once per GPU by a short sweep (\cref{app:tables}).

\subsection{EOQ admission}
\label{sec:eoq}

How many cells should a deep loop accumulate before it advances? Advancing reads one row per request. Waiting delays verification: a backlogged cell may hide a rejection, and every pass of delay is a pass that the request spends on work that will be discarded.

Consider one deep loop with arrival rate $a$ (cells per request per pass) that advances whenever its backlog reaches $B$. Let $\beta$ be the cost of advancing it (\cref{eq:cost}), $\bar\tau$ the mean pass time, $\rho=\beta/\bar\tau$, and $\kappa$ the probability that a cell hides a rejection on this or a deeper loop, taken as independent across cells and stationary over the estimation window. Each cell carries a row cost of $\beta/(NB)$. It waits $B/(2a)$ passes on average, and each pass of delay costs $\kappa\,\bar\tau/N$ in expectation, one wasted request-pass if the cell hides a rejection. Minimizing the sum over $B$ gives
\begin{equation}
B^\star = \sqrt{2a\rho/\kappa},
\label{eq:eoq}
\end{equation}
the economic order quantity of inventory theory, with a row read as the ordering cost and the risk of wasted work as the holding cost (derivation in \cref{app:eoq}). $B^\star$ serves as a real-valued threshold on the backlog. At light load and short context $\rho$ is small and $B^\star<1$: the loop advances whenever it can, which is a wavefront. At heavy load or long context $B^\star$ reaches 2--4, and the loop forms blocks.

Waiting on every deep loop would compound delays. \method{} therefore applies the threshold only to the first deep loop $t_0$ and lets the deeper loops advance whenever they have a backlog. Admitting $t_0$ then implies a row on each of the $T-1$ loops behind it, at total cost $F=(T-1)\beta$, so that
\begin{equation}
B^\star_{t_0} = \sqrt{2aF/(\kappa\bar\tau)},
\label{eq:eoq-lead}
\end{equation}
with $\kappa$ summed over the deep loops. Three of the four inputs are estimated online: $\kappa$ from discard counts, $a$ from the cells each loop computes per pass, and $\bar\tau$ from pass times. $F$ is calibrated once per engine by timing the deep loops' attention at a few context lengths and interpolating (\cref{app:eoq}).

\textbf{Pipeline or blocks.} EOQ prices rows, and accumulation has a second cost. A block passes through the deep loops one loop per pass, so the deep loops advance in different passes and each needs its own attention call. If the deep loops instead advance in every pass, with rows as wide as loop 1's, all $T$ rows of a request share a single call (\cref{sec:engine}). Accumulating therefore pays only once the deep loops' attention is a sufficiently large share of the pass. With $F_n$ the deep loops' attention time in a pipelined pass, comparing the cost per token of the two layouts yields a simple test (\cref{app:eoq}): \method{} pipelines while $F_n/\bar\tau<f^\star=0.4$ and accumulates under \cref{eq:eoq-lead} otherwise. Throughput is insensitive to this constant, which is within 1\% of the best threshold in $[0.2,0.6]$ at 27 of 30 load points. Admission thus needs no per-GPU or per-load table: one configuration moves from a pipeline at light load to blocks at heavy load.

\subsection{Exactness}
\label{sec:exact}

\begin{proposition}
\label{prop:exact}
Under exact arithmetic and greedy decoding, the tokens committed by \method{} equal those of AR decoding with the same model.
\end{proposition}

Call a cell \emph{valid} if its input and all KV entries it attends to were computed from committed or still-valid tokens. Screening discards exactly the cells whose inputs depend on a replaced token, and tokens are committed only from valid loop-$T$ readouts. Scheduling therefore decides only \emph{when} valid cells are computed, never what they contain, so admission, proposal length, and row width cannot change a committed token (proof in \cref{app:proof}).

\textbf{Sampling and early exit.} At temperature $\theta$, every readout returns $\arg\max_v(\ell_v/\theta+g_v)$, with Gumbel noise $g$ fixed by the request and the generation step. All readouts of a position see the same noise, so \cref{prop:exact} carries over: the committed tokens equal those of AR sampling with the same noise and are an exact sample from the model. The argument also covers adaptive early exit, whose decision is a function of valid cells.

\textbf{Floating point.} Batching changes the order of floating-point reductions. In FP32, every \method{} configuration in this paper reproduces AR decoding token for token (\cref{app:tables}). In BF16, AR decoding itself changes its output when the batch shape changes, so we follow the exactness grades of \citet{exactloop} and report task accuracy, which is preserved: on 320 GSM8K problems, AR, wavefront, block, and admission-gated schedules all score 0.7875, and on HumanEval, AR, the screened wavefront, and \method{} pass 75.0\%, 76.2\%, and 75.6\%. \method{} allocates no KV memory beyond AR decoding, since backlogged and discarded cells use positions that the cache already reserves.

\subsection{Implementation}
\label{sec:engine}

All schedules, \method{} and baselines alike, run in one engine with a dense KV cache per loop. Three optimizations apply to every schedule: the shared block is fused and its linear layers compute only the selected cells (\emph{packing}), rows of adjacent loops with equal width share one attention call, and on the RTX 3090 a dedicated kernel serves narrow matrix multiplications. None changes which cells are computed (ablations in \cref{app:tables}).

\section{Experiments}
\label{sec:exp}

The experiments answer four questions: whether \method{} is faster than every existing schedule under load (\textbf{Q1}) and across models, hardware, workloads, and decoding modes (\textbf{Q2}; \cref{sec:main}); whether the lead holds against external systems on their own protocols (\textbf{Q3}; \cref{sec:external}); and whether the gain arises where the cost model of \cref{sec:cost} predicts (\textbf{Q4}; \cref{sec:regime}). \Cref{sec:huginn} turns to deeper recurrence.

\begin{figure*}[t]
\centering
\input{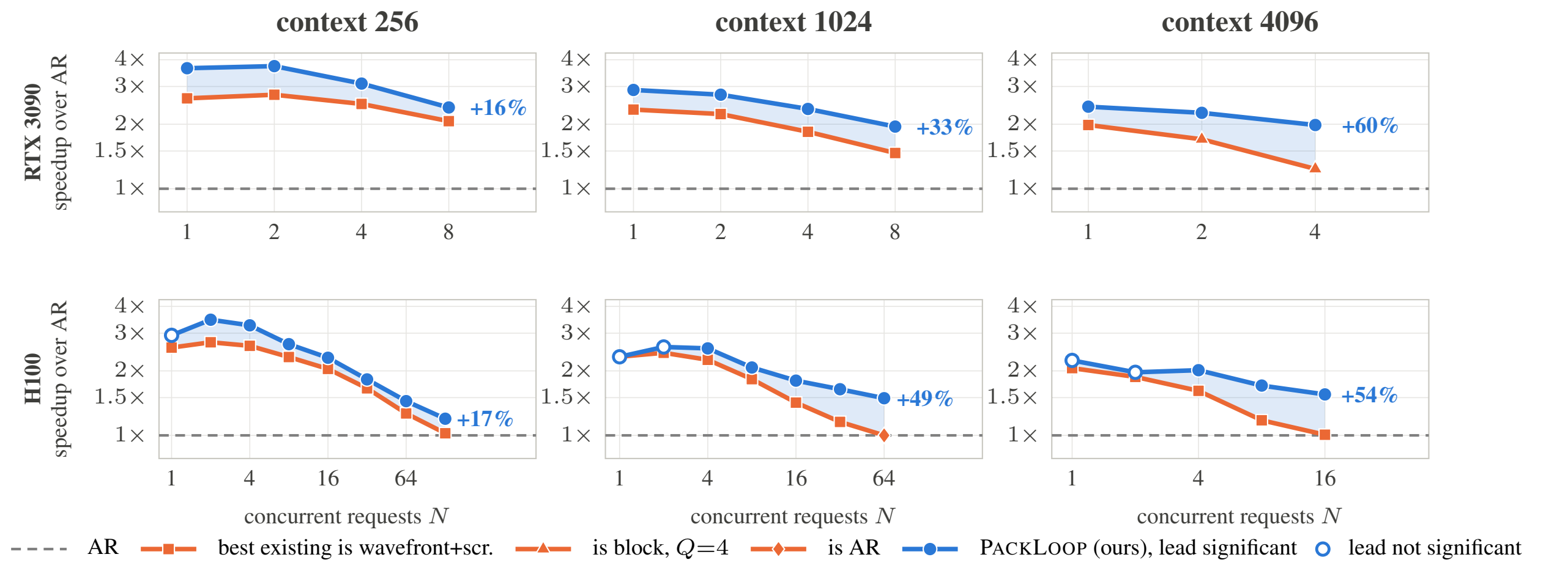}
\begin{tikzpicture}
\pgfplotsset{
  mainfig/.style={
    height=3.3cm, width=0.33\textwidth,
    tick label style={font=\scriptsize, color=cInk},
    label style={font=\scriptsize, color=cInk},
    title style={font=\small\bfseries, color=cInk, yshift=-3pt},
    grid=major, grid style={color=cDone!45, line width=0.3pt},
    axis line style={color=cDone}, major tick length=0pt, minor tick length=0pt, xtick pos=bottom, ytick pos=left, clip=false,
    xmode=log, log basis x=2, ymode=log, ymin=0.78, ymax=4.3, ytick={1,1.5,2,3,4}, yticklabels={$1\times$,$1.5\times$,$2\times$,$3\times$,$4\times$},
    every axis plot/.append style={line width=1.4pt},
  },
  gap/.style={fill=cPL, fill opacity=0.15, draw=none, forget plot},
  arl/.style={gray, densely dashed, line width=0.8pt},
  pl/.style={color=cPL, mark=none, forget plot},
  plsig/.style={only marks, mark=*, mark size=2.1pt, color=cPL, mark options={draw=white, line width=0.4pt}, forget plot},
  plns/.style={only marks, mark=*, mark size=1.9pt, color=cPL, mark options={fill=white, draw=cPL, line width=0.9pt}, forget plot},
  be/.style={color=cWS, mark=none, forget plot},
  bemk/.style={only marks, color=cWS, mark options={draw=white, line width=0.4pt}, forget plot},
  beW/.style={bemk, mark=square*, mark size=1.8pt},
  beB/.style={bemk, mark=triangle*, mark size=2.5pt},
  beA/.style={bemk, mark=diamond*, mark size=2.5pt},
}
\tikzset{lead/.style={font=\scriptsize\bfseries, text=cPL}}
\newcommand{\FMpanel}[3]{%
  \addplot[gap] coordinates {\csname FMfill#1#2\endcsname} -- cycle;
  \addplot[arl, forget plot] coordinates {(0.8,1) (#3,1)};
  \addplot[be] coordinates {\csname FMbe#1#2\endcsname};
  \addplot[beW] coordinates {\csname FMbeW#1#2\endcsname};
  \addplot[beB] coordinates {\csname FMbeB#1#2\endcsname};
  \addplot[beA] coordinates {\csname FMbeA#1#2\endcsname};
  \addplot[pl] coordinates {\csname FMpl#1#2\endcsname};
  \addplot[plsig] coordinates {\csname FMsig#1#2\endcsname};
  \addplot[plns] coordinates {\csname FMns#1#2\endcsname};
  \csname FMlead#1#2\endcsname}
\begin{groupplot}[group style={group size=3 by 2, horizontal sep=0.75cm, vertical sep=0.95cm}, mainfig]
\nextgroupplot[title={context 256}, xmin=0.8, xmax=16, xtick={1,2,4,8}, xticklabels={1,2,4,8}, ylabel={\textbf{RTX 3090}\\speedup over AR}, ylabel style={align=center}]
\FMpanel{T}{a}{16}
\nextgroupplot[title={context 1024}, xmin=0.8, xmax=16, xtick={1,2,4,8}, xticklabels={1,2,4,8}]
\FMpanel{T}{b}{16}
\nextgroupplot[title={context 4096}, xmin=0.8, xmax=8, xtick={1,2,4}, xticklabels={1,2,4}]
\FMpanel{T}{c}{8}
\nextgroupplot[xmin=0.8, xmax=640, xtick={1,4,16,64}, xticklabels={1,4,16,64}, ylabel={\textbf{H100}\\speedup over AR}, ylabel style={align=center}, xlabel={concurrent requests $N$}]
\FMpanel{H}{a}{640}
\nextgroupplot[xmin=0.8, xmax=300, xtick={1,4,16,64}, xticklabels={1,4,16,64}, xlabel={concurrent requests $N$}]
\FMpanel{H}{b}{300}
\nextgroupplot[xmin=0.8, xmax=50, xtick={1,4,16}, xticklabels={1,4,16}, xlabel={concurrent requests $N$},
  legend style={font=\scriptsize, draw=none, fill=none, at={(-0.75,-0.42)}, anchor=north, legend columns=6, column sep=4pt}, legend cell align=left]
\FMpanel{H}{c}{50}
\addlegendimage{arl} \addlegendentry{AR}
\addlegendimage{color=cWS, mark=square*, mark size=1.8pt, mark options={draw=white, line width=0.4pt}} \addlegendentry{best existing is wavefront+scr.}
\addlegendimage{color=cWS, mark=triangle*, mark size=2.5pt, mark options={draw=white, line width=0.4pt}} \addlegendentry{is block, $Q{=}4$}
\addlegendimage{color=cWS, mark=diamond*, mark size=2.5pt, mark options={draw=white, line width=0.4pt}} \addlegendentry{is AR}
\addlegendimage{color=cPL, mark=*, mark size=2.1pt, mark options={draw=white, line width=0.4pt}} \addlegendentry{\method{} (ours), lead significant}
\addlegendimage{only marks, mark=*, mark size=1.9pt, color=cPL, mark options={fill=white, draw=cPL, line width=0.9pt}} \addlegendentry{lead not significant}
\end{groupplot}
\end{tikzpicture}
\caption{Speedup over AR (log scale) of Ouro-1.4B against concurrency at three context lengths on two GPUs; all schedules run in the same engine and job. Orange: the fastest existing schedule at each point (the marker shows which one). Shading: \method{}'s lead over it, labeled at the highest load. Filled circles mark significant leads (paired 95\% interval). \Cref{tab:main} lists all throughputs.}
\label{fig:main}
\end{figure*}

\subsection{Setup}

\textbf{Models and hardware.} Ouro-1.4B, Ouro-2.6B, and their reasoning-tuned Thinking versions ($T=4$; \citealp{ouro}), in BF16 with greedy decoding unless stated, on an NVIDIA RTX 3090 (24\,GB), an H100 NVL (94\,GB), and, for one grid, an RTX 4090 (24\,GB).

\textbf{Workload.} Closed loop with $N\in\{1,\dots,64\}$ concurrent requests, context length $P\in\{256,1024,4096\}$, and 128 generated tokens per request. Prompts come from GSM8K; each load point aggregates several disjoint prompt sets (8 on the RTX 3090 and 4090, 4 on the H100), and all systems receive identical token sequences. We report steady-state throughput (tokens/s); points that do not fit in memory are omitted.

\textbf{Baselines in our engine.} AR; the wavefront with and without screening \citep{wavefront,loopspec}; and blocks with $Q\in\{2,4\}$ and one row per loop. They share the engine, kernels, and job with \method{}. At each load point, \emph{best existing} denotes the fastest of them, an oracle choice in favor of the baselines, whereas \method{} uses one configuration everywhere. Comparing schedules within one engine makes the ratios independent of the speed of the AR baseline, and our AR path is itself 1.2--2.9$\times$ faster than vllm-rlt's no-speculation mode on the same GPU. We pair the two schedules on each prompt set and call a lead \emph{significant} when the 95\% $t$-interval of the throughput ratio excludes 1.

\textbf{External baselines.} vllm-rlt v0.2.0 \citep{vllmrlt} in its fastest mode at each point; continuous depth batching (CDB; \citealp{cdb}) as implemented in vllm-rlt, with exit threshold 0.2; and CopyLoop, WaveFront, and LoopSpec with their released code. LoopAhead has no public code.

\subsection{Main results}
\label{sec:main}

\textbf{Q1: Faster than every existing schedule.} \Cref{fig:main} shows Ouro-1.4B on the two main GPUs. On the RTX 3090, \method{} is 1.16--1.60$\times$ faster than the best existing schedule at all 11 load points, and every lead is significant. On the H100 it leads at 18 of 19 points, by up to 1.54$\times$ and significantly at 14, and it ties the screened wavefront at the remaining point, a single request at context 1024. The lead follows the cost model: it widens with concurrency and with context length, where reads dominate the pass. At 64 requests and context 1024, neither the wavefront nor blocks pay off, so the best existing schedule is AR itself, and \method{} is 1.49$\times$ faster.

\textbf{Q2: The lead generalizes.} We repeat the comparison at 134 further load points: a third GPU, a larger model, reasoning-tuned models in thinking mode, Spec-Bench \citep{specbench} and four more tasks, longer outputs and contexts, higher concurrency, and sampling (\cref{fig:teaser}c; \cref{tab:robust,fig:coverage} in \cref{app:tables}). \method{} is faster than the best existing schedule at 132 of them, significantly at 102, and never significantly slower. Together with the main grid this gives 161 of 164 points, with a median lead of 1.25$\times$. The pattern of the main grid recurs: on the Thinking models the lead is 1.06--1.32$\times$ with 1--8 requests and 1.31--1.48$\times$ with 16--32, and at 8K context it rises from 1.05$\times$ with one request to 1.76$\times$ with eight (1.34$\times$ with two requests at 32K). The lead is smallest on UltraChat (1.03--1.19$\times$), which offers little to copy, and proposals help most on long outputs (4.9$\times$ over AR for one request). Sampling preserves task accuracy: GSM8K scores 66.3, 68.4, and 67.2\% for AR, the screened wavefront, and \method{}.

\textbf{Latency under open-loop arrivals.} Throughput gains carry over to latency. With Poisson arrivals and mixed GSM8K, MBPP, and HumanEval prompts decoded to their natural end, \method{} has the lowest median time per output token and the lowest end-to-end latency at every arrival rate on both GPUs (\cref{fig:openloop}). At the highest rate, its mean end-to-end latency is 3.3\,s on the RTX 3090, against 6.1\,s for the screened wavefront and 35.5\,s for AR, and 1.8\,s on the H100, against 4.0\,s and 43.7\,s.

\subsection{External methods on their own protocols}
\label{sec:external}

\begin{table}[t]
\caption{\ours{} against external systems (throughput ratios). Released decoders use their own engine and inputs (GSM8K / HumanEval); decoders without released code are re-implemented in our engine (54 load points; range, median).}
\label{tab:external}
\centering\small
\setlength{\tabcolsep}{3pt}
\resizebox{\linewidth}{!}{%
\begin{tabular}{llc}
\toprule
System & Setting & \ours{} is faster by \\
\midrule
\multicolumn{3}{l}{\emph{Released decoders, one request}} \\
CopyLoop & RTX 3090 & 1.15$\times$ / 1.20$\times$ \\
LoopSpec, best mode & RTX 3090 & 1.47$\times$ / 2.29$\times$ \\
WaveFront, acceptance 1 & contexts 256--4096 & 2.1--4.2$\times$ \\
\midrule
\multicolumn{3}{l}{\emph{Serving engine vllm-rlt, concurrent requests}} \\
Fastest mode per point & RTX 3090 / H100 & 1.74--2.31$\times$ / 2.18--3.42$\times$ \\
CDB early exit & RTX 3090 / H100 & 1.89--2.60$\times$ / 2.73--3.59$\times$ \\
\midrule
\multicolumn{3}{l}{\emph{Concurrent decoders, re-implemented in our engine}} \\
LoopSkim-style & 3090 and H100 & 0.99--1.39$\times$ (1.07$\times$) \\
LoopFill layout & 3090 and H100 & 1.21--3.29$\times$ (1.78$\times$) \\
\bottomrule
\end{tabular}}
\end{table}

\Cref{tab:external} addresses Q3 by comparing \method{} with each external system under the conditions it was designed for.

\textbf{Single-request decoders.} On CopyLoop's own protocol, \method{} is 1.15--1.20$\times$ faster than CopyLoop and 1.47--2.29$\times$ faster than LoopSpec's best mode, and it outruns WaveFront's released benchmark even at acceptance rate 1, WaveFront's upper bound. In-row proposals thus make \method{} faster than every released decoder in the single-request regime for which they were designed (absolute throughputs in \cref{app:tables}).

\textbf{Serving engine and early exit.} End to end, \method{} is 1.74--3.42$\times$ faster than the fastest vllm-rlt mode at each point; this ratio combines engine and schedule, whereas the lead over the best existing schedule isolates the schedule. Under CDB's exit rule, \method{}'s tokens and exit loops match a cell-by-cell reference, and with identical inputs it is 1.89--3.59$\times$ faster than CDB in vllm-rlt. To our knowledge, \method{} is the first looped-model decoder in which exact speculation and adaptive early exit operate together \citep{cdb,vllmrlt}.

\textbf{Concurrent decoders.} LoopSkim and LoopFill have no released code, so we re-implement both in our engine, in the jobs of the main grids (54 points). The LoopSkim-style schedule copies drafts on suffix matches of at least two tokens, checks them at every loop, and pipelines them; as LoopSkim tunes per batch size, it takes the faster of 2 and 4 draft tokens at each point, and we grant it our row-sharing layout. \method{} remains faster, by 1.07$\times$ at the median, significantly at 43 points and never significantly slower, and its lead reaches 1.10--1.39$\times$ at the 12 read-heavy points ($N P \ge 16$K), where it also packs the deep loops. With the same drafts in LoopFill's layout, in which every draft position occupies a row at every loop, \method{} is 1.78$\times$ faster at the median (\cref{tab:concurrent}). Given LoopSkim's trained draft head as well (\cref{app:dhead}), one \method{} configuration matches LoopSkim's best setting per point (median 0.98$\times$) and is up to 1.18$\times$ faster where reads dominate.

\subsection{Where the gain comes from}
\label{sec:regime}

\textbf{Q4: Two sources of gain, selected by the cost model.} The layout chosen by the admission rule shows which source of gain is active (\cref{fig:rho}). At 18 of the 30 Ouro-1.4B points, those with light load or short context, \method{} runs the deep loops as a pipeline. Extra cells are nearly free there, and in-row proposals carry the gain, 1.15$\times$ at the median. At the other 12 points reads dominate, EOQ admission accumulates blocks of 2--5 cells, and row sharing carries the gain, 1.33$\times$ at the median and up to 1.60$\times$. Across the 30 points the gain rises with $\rho$, the share of a pass spent reading one loop's rows.

\textbf{The cost model is predictive.} A replay of the engine's state machine and admission rule on recorded per-loop trajectories, with pass costs fitted only to the microbenchmark of \cref{fig:teaser}(a), predicts the speedups over AR at the RTX 3090 points within 4.1\% at the median and ranks \method{} first at every point (\cref{app:huginn}).

\begin{table}[t]
\caption{Ablations on Ouro-1.4B (RTX 3090, 11 load points): throughput relative to \method{} in the same job when one part is changed.}
\label{tab:ablation}
\centering\small
\resizebox{\linewidth}{!}{%
\begin{tabular}{lcc}
\toprule
Variant & $\times$\method{} & Worst point ($P$, $N$) \\
\midrule
\oursrow \ours{} & 1.00 & -- \\
\midrule
\multicolumn{3}{l}{\emph{Same speculative work, different layout}} \\
One row per position & 0.18--0.66 & 4096, 2 \\
One row per proposal position (CopyLoop layout) & 0.38--0.79 & 4096, 2 \\
Proposals verified round by round on AR & 0.45--0.65 & 256, 1 \\
\midrule
\multicolumn{3}{l}{\emph{Components of \method{}}} \\
Without in-row proposals & 0.72--0.88 & 256, 1 \\
Synchronized blocks $Q{=}4$, no proposals or admission & 0.59--0.92 & 256, 1 \\
Fixed cadence $B{=}2$ at all points & 0.89--1.02 & 256, 4 \\
Fixed cadence, best per point (oracle) & 0.91--1.02 & 256, 4 \\
Without the pipeline test (always accumulate) & 0.92--1.01 & 256, 4 \\
Without load-set width (loop-1 width 8) & 0.77--1.00 & 256, 8 \\
\bottomrule
\end{tabular}}
\end{table}

\textbf{Every component contributes.} \Cref{tab:ablation} changes one part of \method{} at a time. \emph{Layout} matters most: the same speculative work with one row per position, one row per proposal position, or round-by-round verification retains only 0.18--0.79$\times$ of \method{}'s throughput. \emph{Proposals} carry the gain at light load: without them \method{} keeps 72--80\% of its throughput with one request and 86--88\% with eight. \emph{Admission} removes tuning: an oracle that picks the best fixed cadence at every point is at most 2\% faster than \method{} and up to 9\% slower, and the best single cadence loses up to 11\%. The \emph{pipeline test} and the \emph{load-set width} protect the short-context points, where always accumulating loses up to 8\% and 8-cell loop-1 rows lose 16--23\% at 8 requests.

\subsection{Deeper recurrence}
\label{sec:huginn}

\textbf{Measured, $T=8$.} Running Ouro-1.4B with eight iterations on the H100, \method{} is 1.4--4.5$\times$ faster than AR at every point and 1.21--1.91$\times$ faster than the best existing schedule at the five read-heavy points.

\textbf{Predicted, $T=32$.} Huginn-0125 \citep{huginn} iterates a 4-layer block 32 times. We predict \method{} on it with the replay validated above, from per-iteration trajectories of 120 requests and the measured cost of the recurrent block (\cref{app:huginn}). Shallow readouts are less reliable at this depth, so drafts come from iteration 8, and in-row proposals let one traversal of this draft chain advance several positions. With one configuration for all points, the predicted lead over the best existing schedule has a median of 1.19$\times$ over 25 points on both GPUs and reaches 1.35$\times$.

\section{Related Work}
\label{sec:related}

\textbf{Looped models and their decoders.} Universal Transformers \citep{universal} share weights across depth, and Huginn \citep{huginn} and Ouro \citep{ouro} scale the idea to language models with billions of parameters; FlashLoop \citep{flashloop} reduces their KV memory and prefill cost. SPEED \citep{speed} first pipelined speculative tokens through parameter-shared decoder layers. WaveFront, LoopSpec, LoopAhead, and CopyLoop \citep{wavefront,loopspec,loopahead,copyloop} use shallow loops or the context as drafters and pipeline positions across loops; all are designed and evaluated for one request, and their cost models count passes. Two concurrent decoders evaluate batches. LoopFill \citep{loopfill} combines suffix drafts \citep{suffixdecoding} and shallow-loop candidates in a draft tree in which every node occupies a row at every loop. LoopSkim \citep{loopskim} pipelines copied drafts that are checked at every loop and tunes the draft length per batch size. Neither models the cost of a pass. \method{} keeps the pipeline, makes positions share rows within it, and derives how much to speculate from the measured cost.

\textbf{Serving looped models.} vllm-rlt \citep{vllmrlt} provides paged KV, early exit, and block and pipeline speculation with one row per verified token. \citet{exactloop} find that fixed-depth self-speculation on vllm-rlt turns into a slowdown at 32 requests and attribute the loss to a GPU saturated by the target computation. We locate the cost in rows of KV reads and show that speculation stays profitable at high load once rows are shared. Continuous depth batching \citep{cdb} lets tokens exit at their own depth, without speculation, and vllm-rlt's speculative path requires fixed depth; \method{} supports both, exactly.

\textbf{Load-aware speculation and shared reads.} TurboSpec \citep{turbospec}, Nightjar \citep{nightjar}, AdaSpec \citep{adaspec}, and FASER \citep{faser} adapt the speculation length to load, typically speculating less at high load; ECHO \citep{echo} and D-cut \citep{dcut} share a budget of verification tokens across the batch, and \citet{exspec} show how batching can break output equivalence. These systems ration verification compute; in a looped model the scarce resource is rows of KV reads, so \method{} keeps speculating at every load and changes where the work is placed. Hydragen \citep{hydragen} reads a prefix shared by many sequences once; \method{} shares reads among positions of one request on one loop. In non-looped self-speculation \citep{draftverify,layerskip} each layer has its own cache, so the row problem is specific to pipelining over a looped block.

\section{Discussion and Conclusion}
\label{sec:limits}

\textbf{Scope and extensions.} We measure the Ouro family at $T=4$ and $8$ with a dense KV cache and predict Huginn at $T=32$ with a validated replay. At large $T$ most rows belong to the draft chain, so a faster drafter such as LoopAhead's compounds with packing. Rows and admission are independent of the acceptance test, so LoopSpec's cascaded rejection sampling can replace the fixed-noise readout. With a paged cache, a row is one pass over a request's pages.

\textbf{Conclusion.} Under concurrency, the cost of decoding a looped model is counted in rows of KV reads, not in passes, and pipelined self-speculation loses its gains because it gives every speculative position a row of its own. \method{} keeps the pipeline and makes positions share rows under a closed-form admission rule, with exact outputs. With one configuration it outperforms the per-point best existing schedule at 161 of 164 load points, by up to 1.76$\times$, and most where existing methods fall back to AR decoding. More broadly, speculative decoders for weight-shared models should be designed around cache reads, not sequential passes.

\section*{Code Availability}
Code, scripts, and logs will be released.

\section*{Impact Statement}
This paper presents work whose goal is to advance the field of machine learning, specifically the efficiency of inference for looped language models. Faster exact decoding reduces the energy and hardware needed to serve such models. We do not see societal consequences specific to this work beyond those of language models in general.

\bibliography{refs}
\bibliographystyle{icml2026}

\newpage
\appendix
\onecolumn

\section{Notation}
\label{app:notation}
\begin{tabular}{ll}
$T$ & number of loops; $t$ indexes loops \\
$N$, $P$ & concurrent requests; context length \\
row & one request advancing on one loop in a pass (one KV read) \\
cell & one position computed on one loop \\
$Q$ & maximum cells a deep loop advances in one row ($Q=8$) \\
$m$ & maximum proposal length ($m=8$) \\
$B$, $B^\star$ & backlog threshold for admitting a deep loop; its optimum \\
$c_0$, $\beta$, $\gamma$, $g$ & fixed pass cost; cost of one loop's rows; scale and calibrated shape of cell cost \\
$\bar\tau$, $\rho$ & mean pass time; $\beta/\bar\tau$ \\
$a$, $\kappa$ & cells arriving per request per pass; probability that a cell hides a rejection \\
$F$, $F_n$, $f^\star$ & deep loops' attention cost (blocks; pipelined pass); pipeline-test threshold \\
$\theta$ & sampling temperature \\
$w_n$ & load threshold for the loop-1 row width \\
$K$ & draft length in vllm-rlt's block mode \\
\end{tabular}

\section{Proof of Exactness}
\label{app:proof}
\textbf{State.} For one request, let $x_1,\dots,x_c$ be the committed tokens and $y_{c+1},y_{c+2},\dots$ the current hypothesis: drafted tokens and proposals not yet committed. The input token of position $i$ is $u_i = x_i$ if $i\le c$ and $u_i = y_i$ otherwise. For each loop $t$ the scheduler keeps a frontier $f_t$, the number of positions whose cell on loop $t$ has been computed and not discarded, with $f_1\ge f_2\ge\dots\ge f_T$. Cell $(i,t)$ stores $h^{(t)}_i$ and its keys and values.

\textbf{Invariant.} For every loop $t$ and every $i\le f_t$, the stored $h^{(t)}_i$ and $\mathrm{KV}^{(t)}_i$ equal the values that AR decoding would compute if its tokens were $u_1,\dots,u_i$. Call such a cell \emph{consistent}.

\emph{Advancing preserves the invariant.} A pass advances loop $t$ from $f_t$ to $f_t+r$ only if $f_t+r\le f_{t-1}$ (and, for $t=1$, only over positions whose input token is known). Cell $(i,t)$ is computed from $h^{(t-1)}_i$, which is consistent, and attends to $\mathrm{KV}^{(t)}_{j}$ for $j\le i$: the entries $j\le f_t$ are consistent, and those with $f_t<j\le i$ are computed in the same row of the same pass under the causal mask, from consistent inputs. The computation is the same function as in AR decoding, so the new cells are consistent with respect to $u$. Packing, grouping of loops, and padding change only which slots of the dense layout are computed; padded slots are never read as valid cells.

\emph{Replacing a hypothesis token preserves the invariant.} When a readout at position $i$ on loop $t$ disagrees with $u_{i+1}$, the hypothesis is replaced from position $i{+}1$ on, and every frontier is truncated to at most $i{+}1$. Cells at positions $\le i{+}1$ depend only on $u_1,\dots,u_{i+1}$; the input $u_{i+1}$ itself only enters loop 1 at position $i{+}1$, which is recomputed. So all remaining cells are consistent with respect to the new hypothesis.

\emph{Commitment.} A token is committed only from the loop-$T$ readout at position $i=c$, which by the invariant equals the AR readout given $x_1,\dots,x_c$. If it disagrees with $y_{c+1}$, the hypothesis is replaced as above before the next pass. Hence by induction $x_{c+1}$ equals the $(c{+}1)$-th AR token. Admission (\cref{sec:eoq}), proposal length, and row width only decide \emph{which} consistent cells are computed in a pass, so they cannot change the committed tokens.

\textbf{Early exit.} With an exit rule, the readout used to commit position $i$ is the one of the exit loop $e_i$, and loops deeper than $e_i$ reuse the KV of loop $e_i$ at that position. The exit decision is a function of the gates of loops $\le e_i$ at position $i$, all of which are consistent cells, so the reference implementation that evaluates the rule cell by cell makes the same decisions. We verified this on GPU: under CDB's rule, tokens and per-token exit loops match the reference on all requests checked (851--956 tokens per policy, FP32).

\textbf{Floating point.} The argument assumes that each cell is a deterministic function of its inputs. In FP32 with TF32 disabled this holds closely enough that every configuration in this paper produced the same tokens as AR decoding (\cref{app:tables}). In BF16, the reduction order of batched kernels depends on the batch shape, so near-ties of the argmax can flip; AR decoding itself changes its output when one request is added to the batch, with the first divergence at a top-1/top-2 logit gap of one BF16 step (0.125 in median).

\section{EOQ Admission: Derivation and Estimators}
\label{app:eoq}

\textbf{Derivation.} Consider one deep loop in steady state. Each pass, $a$ cells per request arrive in its backlog. If the loop advances when the backlog reaches $B$, it advances every $B/a$ passes, and each advance costs $\beta$ (one row per request, shared by the $N$ requests). The row cost per cell is $\beta/(NB)$. A cell arriving in the backlog waits on average $B/(2a)$ passes. If it hides a rejection (probability $\kappa$), every pass of delay is a pass in which its request computes cells that will be discarded; in time, that is one request-pass, $\bar\tau/N$. The expected delay cost per cell is $\kappa\,\bar\tau B/(2aN)$. The total per cell,
\[
C(B) = \frac{\beta}{NB} + \frac{\kappa\,\bar\tau\,B}{2aN},
\]
is minimized at $B^\star=\sqrt{2a\beta/(\kappa\bar\tau)}=\sqrt{2a\rho/\kappa}$. This is the economic order quantity with ordering cost $\beta$, demand rate $a$, and holding cost $\kappa\bar\tau$ per cell and pass. When only the first deep loop $t_0$ is thresholded, admitting it implies one row on each of the $T-1$ deep loops (the deeper loops advance whenever they have a backlog), so the ordering cost is $(T-1)\beta$; a cell then waits only once, and $\kappa$ is the probability of a rejection on any deep loop, giving \cref{eq:eoq-lead}.

\textbf{Assumptions.} The derivation treats rejections as independent across cells and the rates as stationary over the estimation window, and it ignores the integrality of $B$: the backlog is compared with the real-valued threshold each pass. It also ignores the effect of the threshold on $\bar\tau$ itself; the estimates are updated every pass, so the threshold tracks the resulting pass time.

\textbf{Online estimates.} All estimates use exponential forgetting so they follow changes in load.
\begin{itemize}
\item $a$: an exponential moving average (weight 0.03) of the cells the loop computes per request per pass. In steady state arrivals equal departures; using the upstream loop's cells would count proposal cells that loop 1 discards on the spot.
\item $\kappa$: for each deep loop, discards over computed cells with decay 0.998 and a prior of 30 cells at rate 0.1, summed from the loop to the deepest one.
\item $F$: timed once per engine by replaying the deep loops' attention, summed over the block's layers, in a CUDA graph (20 repetitions), for both layouts: $F_n$ is the time of one call over all $T$ loops at loop 1's width minus that of loop 1 alone, and $F_w$ the time of the deep loops at width $Q$. In closed-loop runs it is timed at the shortest and longest context of the run and interpolated linearly; when serving, request contexts differ, so it is timed at seven contexts from 128 up to the cache size and interpolated piecewise linearly. Only cached keys and values are read, so the decoding state is untouched.
\item $\bar\tau$: an exponential moving average (weight 0.05) of pass times, in the layout in use.
\end{itemize}
The pipeline test uses $F_n/\bar\tau$; the block threshold uses $F_w$ in \cref{eq:eoq-lead}.
Two safeguards keep the pipeline moving: if no request can draft (all hypotheses are at their length limit), every loop with a backlog is admitted; and a backlog that has waited more than $2\max(B^\star,1)/a$ passes is admitted.

\textbf{Why $F$ is calibrated rather than regressed.} The pipeline test needs the cost of both layouts, and pass times show only the layout in use. Fitting $\beta$ online from pass times with \cref{eq:cost} is also fragile. Because only $t_0$ is thresholded, the deep loops advance together, so the number of advancing loops is almost always 1 or $T$ and changes together with the number of cells. Every regression we tried attributes part of one cost to the other:
\begin{itemize}
\item \emph{Unconstrained regression} on $(1, P_{\text{act}}, C)$: when a deep loop is admitted, both regressors jump, and the steps of the narrow GEMMs are attributed to $\beta$. For one HumanEval request, $\rho$ is estimated at 0.08 against at most 0.02 from paired measurements; blocks become too large and passes per token rise by 12\%.
\item \emph{Subtracting the calibrated $g$} and regressing the residual on $(1, P_{\text{act}})$: correct when $g$ has the right scale, but under high load on the H100 the scale is off and $\beta$ is pushed to zero (throughput $-14\%$ at $P=256$, $N=32$ in one run; $-4.6\%$ at $P=4096$, $N=1$).
\item \emph{Fixed effects by cell count}, with one intercept per cell count: robust to the shape of $g$ but leaks per-cell overheads not captured by the bins ($\rho$ 0.033 against 0.018).
\item \emph{Calibrated shape with fitted scale}: regress on $(1, P_{\text{act}}, g(C))$. The steps of $g$ are explained by $g$, so they do not leak into $\beta$, and a wrong scale is absorbed by $\gamma$. It matches the subtraction estimator within $-1.3\%$ to $+1.1\%$ on the RTX 3090 and removes both H100 losses.
\end{itemize}
Where the rule accumulates, the calibrated threshold and this last regression give the same throughput. Without the pipeline test, admission with the calibrated $F$, with this regression, and with the unconstrained regression runs at 0.92--1.01$\times$, 0.90--1.01$\times$, and 0.89--1.01$\times$ of \method{}'s throughput on the 11 RTX 3090 points (worst points: context 256 with 4, 4, and 8 requests); \cref{tab:ablation} lists the first.

\textbf{The pipeline test.} EOQ prices rows, not attention calls. A block passes through the deep loops one loop per pass, so the deep loops advance in different passes and each needs its own attention call; together these calls cost $r=1.4$--$2.4\times$ one call over the same rows (RTX 3090). If the deep loops instead advance whenever they have a backlog, with rows as wide as loop 1's, all $T$ rows of a request share one call. Blocks save deep rows but add about 7\% passes per token for each pass of their period $p$ beyond the first. With $F_n$ the deep loops' attention in a pipelined pass and $\tau_1$ the rest of the pass, a pipeline costs $\tau_1+F_n$ per pass and blocks cost $(1+0.07(p-1))(\tau_1+rF_n/p)$. The two are equal at $F_n/(\tau_1+F_n)\approx0.37$ for $p=3$ and $r=2$, which gives the threshold $f^\star=0.4$. Since $\bar\tau$ is measured in the layout in use and blocks shorten passes, returning to the pipeline takes a clearly lower load, so the rule does not oscillate between layouts.

\textbf{Admission and the best fixed cadence.} \Cref{tab:bstar} compares the layout and threshold the engine chooses for the first deep loop (at the end of each run) with fixed cadences $B\in\{1,2,3,4\}$ run in the same job (RTX 3090, Ouro-1.4B, 8 prompt sets); fixed cadences always use blocks. At the seven points where \method{} pipelines, $B=1$ is best at five; at context 256 with 4 and 8 requests $B=3$ is best, and the pipeline is 10\% and 3\% faster than it. At the four points where \method{} accumulates, its threshold is 1.8--3.1 and the best cadence is $B=3$ or $4$. At every point \method{} is within 2\% of the best fixed cadence.

\begin{table}[h]
\caption{Layout and EOQ threshold $B^\star_{t_0}$ chosen by \method{} against the best fixed cadence (RTX 3090). Last column: \method{}'s throughput over the best fixed cadence's.}
\label{tab:bstar}
\centering\small
\begin{tabular}{rrlcccc}
\toprule
$P$ & $N$ & layout & $\hat\rho$ & $B^\star_{t_0}$ & best $B$ & ratio \\
\midrule
256 & 1 & pipeline & 0.026 & -- & 1 & 0.99 \\
256 & 2 & pipeline & 0.058 & -- & 1 & 1.02 \\
256 & 4 & pipeline & 0.087 & -- & 3 & 1.10 \\
256 & 8 & pipeline & 0.117 & -- & 3 & 1.03 \\
1024 & 1 & pipeline & 0.056 & -- & 1 & 1.00 \\
1024 & 2 & pipeline & 0.108 & -- & 1 & 0.98 \\
1024 & 4 & blocks & 0.201 & 1.83 & 3 & 0.99 \\
1024 & 8 & blocks & 0.266 & 3.14 & 3 & 1.00 \\
4096 & 1 & pipeline & 0.132 & -- & 1 & 1.00 \\
4096 & 2 & blocks & 0.283 & 2.35 & 4 & 0.99 \\
4096 & 4 & blocks & 0.389 & 2.98 & 4 & 0.98 \\
\bottomrule
\end{tabular}
\end{table}

\textbf{Sensitivity to $f^\star$.} We reran the Ouro-1.4B grid with $f^\star\in\{0.2,0.3,0.4,0.5,0.6\}$, all five in the same job (RTX 3090: 11 points, 8 prompt sets; H100: 19 points, 4 sets). The derived $f^\star=0.4$ is within 1\% of the best of the five at 27 of the 30 points and within 4.5\% at all of them, and 0.3 behaves almost the same. Moving the threshold further changes the layout at points near the boundary: 0.2 starts accumulating too early (up to $-12\%$) and 0.6 keeps pipelining too long (up to $-20\%$).

\section{Deeper Recurrence: Replay Details}
\label{app:huginn}

\begin{figure}[h]
\centering
\input{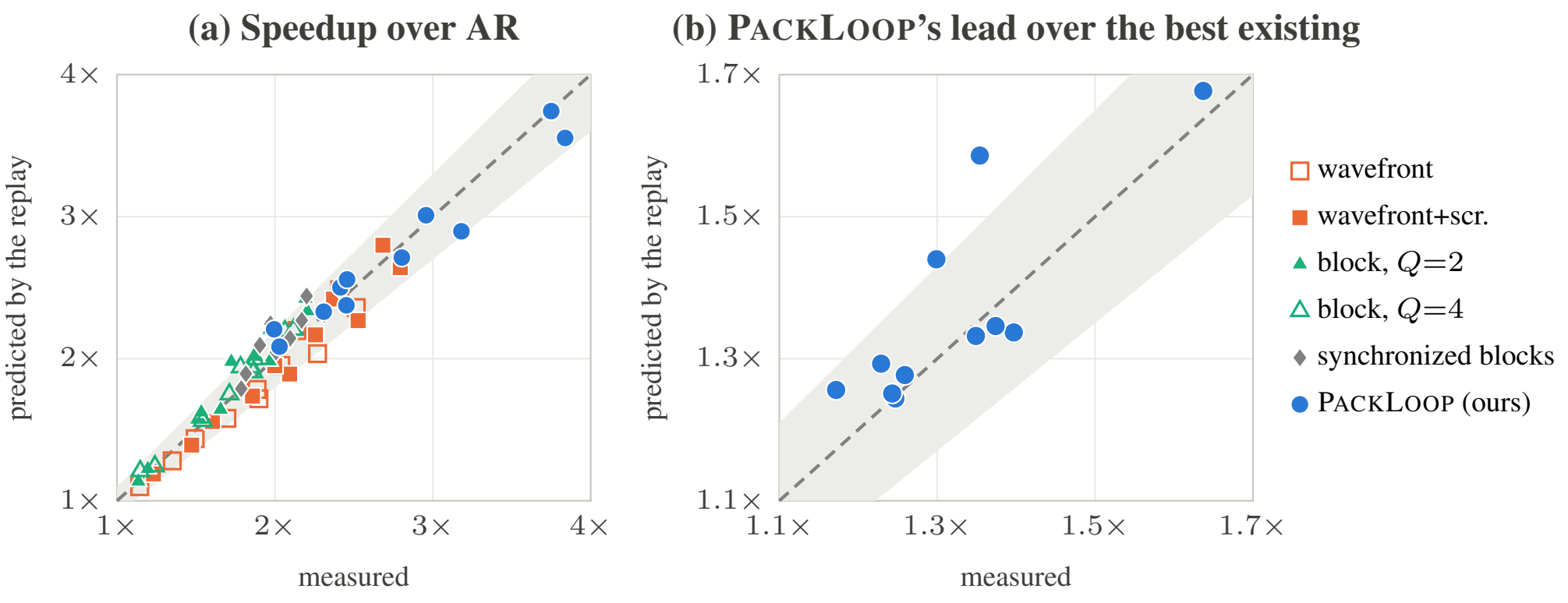}
\begin{tikzpicture}
\pgfplotsset{
  vlfig/.style={
    height=5.2cm, width=5.6cm,
    tick label style={font=\scriptsize, color=cInk},
    label style={font=\scriptsize, color=cInk},
    title style={font=\small\bfseries, color=cInk, yshift=-3pt},
    grid=major, grid style={color=cDone!45, line width=0.3pt},
    axis line style={color=cDone}, major tick length=0pt, minor tick length=0pt, xtick pos=bottom, ytick pos=left, clip=false,
    xlabel={measured}, ylabel={predicted by the replay},
  },
  vmk/.style={only marks, mark size=2pt, mark options={line width=0.6pt}},
}
\begin{groupplot}[group style={group size=2 by 1, horizontal sep=1.6cm}, vlfig]
\nextgroupplot[title={(a) Speedup over AR}, xmin=1, xmax=4, ymin=1, ymax=4, xtick={1,2,3,4}, ytick={1,2,3,4},
  xticklabels={$1\times$,$2\times$,$3\times$,$4\times$}, yticklabels={$1\times$,$2\times$,$3\times$,$4\times$},
  legend style={font=\scriptsize, draw=none, fill=none, at={(2.45,0.5)}, anchor=west, row sep=1pt}, legend cell align=left]
\addplot[fill=cDone!35, draw=none, forget plot] coordinates {(1,1) (1.111,1) (4,3.6) (4,4) (3.636,4) (1,1.1)} -- cycle;
\addplot[gray, densely dashed, line width=0.8pt, forget plot] coordinates {(1,1) (4,4)};
\addplot[vmk, mark=square, color=cWS] coordinates {\VLw}; \addlegendentry{wavefront}
\addplot[vmk, mark=square*, color=cWS, mark options={draw=white}] coordinates {\VLws}; \addlegendentry{wavefront+scr.}
\addplot[vmk, mark=triangle*, color=cBQ, mark size=2.4pt, mark options={draw=white}] coordinates {\VLbb}; \addlegendentry{block, $Q{=}2$}
\addplot[vmk, mark=triangle, color=cBQ, mark size=2.4pt] coordinates {\VLbf}; \addlegendentry{block, $Q{=}4$}
\addplot[vmk, mark=diamond*, color=gray, mark size=2.4pt, mark options={draw=white}] coordinates {\VLsy}; \addlegendentry{synchronized blocks}
\addplot[vmk, mark=*, color=cPL, mark size=2.2pt, mark options={draw=white}] coordinates {\VLpl}; \addlegendentry{\method{} (ours)}
\nextgroupplot[title={(b) \method{}'s lead over the best existing}, xmin=1.1, xmax=1.7, ymin=1.1, ymax=1.7, xtick={1.1,1.3,1.5,1.7}, ytick={1.1,1.3,1.5,1.7},
  xticklabels={$1.1\times$,$1.3\times$,$1.5\times$,$1.7\times$}, yticklabels={$1.1\times$,$1.3\times$,$1.5\times$,$1.7\times$}]
\addplot[fill=cDone!35, draw=none] coordinates {(1.1,1.1) (1.222,1.1) (1.7,1.53) (1.7,1.7) (1.545,1.7) (1.1,1.21)} -- cycle;
\addplot[gray, densely dashed, line width=0.8pt] coordinates {(1.1,1.1) (1.7,1.7)};
\addplot[vmk, mark=*, color=cPL, mark size=2.4pt, mark options={draw=white}] coordinates {\VLlead};
\end{groupplot}
\end{tikzpicture}
\caption{Replay predictions against measurements on the RTX 3090 (Ouro-1.4B, 11 load points; measured throughput on GPU time, as the replay does not model the host). (a) Speedup over AR of every schedule. (b) \method{}'s lead over the per-point fastest existing schedule. Dashed: equality; shaded: within 10\%.}
\label{fig:validate}
\end{figure}

\textbf{Validation on Ouro.} We replay the engine's own state machine and admission rule on per-loop greedy trajectories recorded on the main-table prompts, and charge each pass with an affine form of \cref{eq:cost} fitted only to the row/cell microbenchmark of \cref{fig:teaser}(a); no main-table number enters the fit. Over the 11 RTX 3090 points and 7 schedules, the replay predicts passes per token within 0.6\% (median; 10\% at most), speedups over AR within 4.1\% (median; 10\% at the 90th percentile), and \method{}'s lead over the best existing schedule within 2.5\% (median; \cref{fig:validate}). It ranks \method{} first at every point.

\textbf{Ouro at $T=8$.} The five read-heavy points of the measured $T=8$ comparison (\cref{sec:huginn}) are 64 requests at context 256, 16--32 requests at context 1024, and 4--8 requests at context 4096.

\textbf{Huginn.} Huginn-0125 \citep{huginn} runs a 4-layer block $T=32$ times between a 2-layer prelude and a 2-layer coda. We predict \method{} on it with the replay validated above: per-iteration greedy trajectories of 120 GSM8K and MBPP requests, and the cost of the recurrent block measured on the H100 and the RTX 3090, with the prelude and coda charged as half a block each. Every schedule takes its best draft depth and block length per point, and requests are limited by KV memory (24 at context 1024 on the H100, 4 on the 3090).

Depth changes where the rows are. The readout after one iteration agrees with the final token for only 31--58\% of tokens, and after eight for 92--93\%, so the best draft depth is $d=8$. Drafting then costs eight sequential passes per token, and the draft chain caps each request at one token per eight passes. Each read loop also adds a coda pass, so drafts are screened only at loop $d$. Packing the 24 deep loops cuts rows per token from 36 (wavefront) to about 24, but the chain's eight rows per token remain. In-row proposals lift the cap: with $d>1$, a proposal rides in the rows of loops $1,\dots,d$ and is screened at loop $d$, so one traversal of the chain advances several positions, and passes per token fall from about 10 to about 7.3. Outputs stay exact.

With one configuration for all points ($d=8$, $Q=16$, proposals of 7 tokens for up to two requests and 3 beyond), the predicted speedup of \method{} over AR is 2.9--3.8$\times$ with one or two requests and 1.8--2.1$\times$ with 16--32 requests on the H100, against 2.4--2.8$\times$ and 1.7--2.1$\times$ for the best existing schedule. Over 25 points on both GPUs, the predicted lead has a median of 1.19$\times$ and reaches 1.35$\times$. Huginn's reference speculative decoder (draft, then verify with full recomputation) is predicted at 1.4--1.8$\times$ with per-sequence acceptance, and below AR from 8 requests on as batched, where every sequence rolls back to the shortest accepted prefix.

\section{Further Results}
\label{app:tables}

\begin{figure*}[h]
\centering
\input{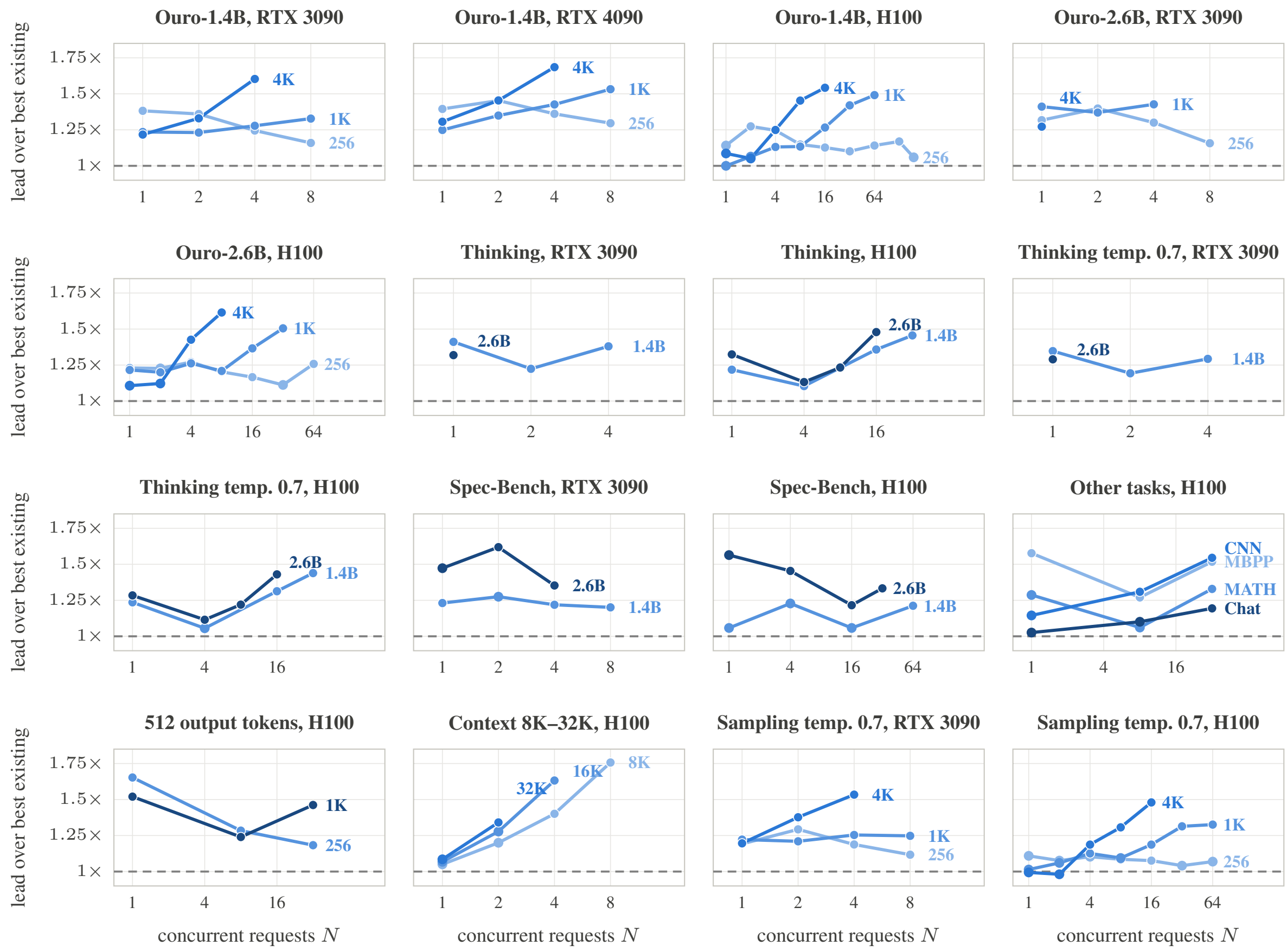}
\caption{\method{}'s lead over the per-point fastest existing schedule in the same engine and job, at every load point, by setting and GPU. One line per context length (or model, or task), labeled at its end. Filled: the paired 95\% interval lies above 1; hollow: it does not. Dashed: parity.}
\label{fig:coverage}
\end{figure*}

\textbf{Every load point.} \Cref{fig:coverage} plots \method{}'s lead over the per-point fastest existing schedule at all 164 load points of \cref{fig:main,tab:robust}. \method{} is faster at 161 of them, significantly at 127, and never significantly slower. Within a setting, the lead grows with concurrency at long context and is largest where rows dominate the pass. \Cref{tab:robust} summarizes the 134 points beyond the main grid by setting and GPU.

\begin{table}[h]
\caption{Beyond the grid of \cref{fig:main}: \method{}'s speedup over AR and lead over the per-point fastest existing schedule (ranges). Sig.: points with a significant lead. Thinking: Ouro-1.4B/2.6B-Thinking, MATH-500, 1024 output tokens. Other tasks: MBPP, MATH-500, CNN/DailyMail, UltraChat.}
\label{tab:robust}
\centering\footnotesize
\setlength{\tabcolsep}{3pt}
\begin{tabular}{llrccr}
\toprule
Setting & GPU & Pts & $\times$AR & vs best & sig. \\
\midrule
Ouro-1.4B & 4090 & 11 & 2.24--4.04 & 1.25--1.68 & 11 \\  %
Ouro-2.6B & 3090 & 8 & 2.48--4.12 & 1.16--1.43 & 8 \\  %
Ouro-2.6B & H100 & 17 & 1.62--3.69 & 1.11--1.62 & 13 \\  %
\midrule
Thinking, greedy & 3090 & 4 & 2.51--3.55 & 1.22--1.41 & 4 \\  %
Thinking, greedy & H100 & 8 & 1.68--3.86 & 1.10--1.48 & 8 \\  %
Thinking, temp.\ 0.7 & 3090 & 4 & 2.29--3.43 & 1.19--1.35 & 4 \\  %
Thinking, temp.\ 0.7 & H100 & 8 & 1.63--3.78 & 1.06--1.44 & 7 \\  %
Spec-Bench & 3090 & 7 & 1.68--3.90 & 1.20--1.62 & 5 \\  %
Spec-Bench & H100 & 8 & 1.21--4.41 & 1.06--1.56 & 4 \\  %
Other tasks & H100 & 12 & 1.29--4.76 & 1.03--1.58 & 7 \\  %
\midrule
512 output tokens & H100 & 6 & 1.72--4.91 & 1.18--1.65 & 6 \\  %
Context 8K--32K & H100 & 9 & 1.51--1.91 & 1.05--1.76 & 4 \\  %
128--192 requests & H100 & 2 & 1.06--1.20 & 1.06--1.17 & 1 \\  %
Sampling, temp.\ 0.7 & 3090 & 11 & 1.74--3.24 & 1.12--1.53 & 11 \\  %
Sampling, temp.\ 0.7 & H100 & 19 & 1.23--2.89 & 0.98--1.48 & 9 \\  %
\bottomrule

\end{tabular}
\end{table}

\begin{figure}[h]
\centering
\input{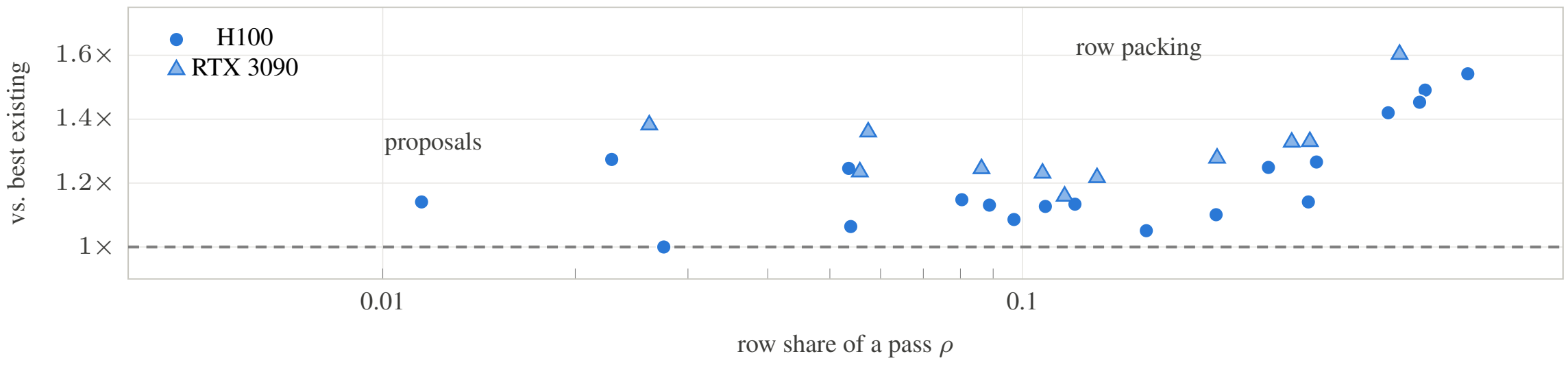}
\begin{tikzpicture}
\begin{axis}[height=4.2cm, width=0.9\columnwidth,
  tick label style={font=\scriptsize, color=cInk}, label style={font=\scriptsize, color=cInk},
  grid=major, grid style={color=cDone!45, line width=0.3pt}, axis line style={color=cDone}, major tick length=0pt, xtick pos=bottom, ytick pos=left, clip=false,
  xmode=log, xmin=0.004, xmax=0.7, xtick={0.01,0.1}, xticklabels={0.01,0.1},
  ymin=0.9, ymax=1.75, ytick={1,1.2,1.4,1.6}, yticklabels={$1\times$,$1.2\times$,$1.4\times$,$1.6\times$},
  xlabel={row share of a pass $\rho$}, ylabel={vs.\ best existing},
  legend style={font=\scriptsize, draw=none, fill=none, at={(0.02,0.98)}, anchor=north west, row sep=-2pt},
]
\addplot[gray, densely dashed, line width=0.8pt, mark=none, forget plot] coordinates {(0.004,1) (0.7,1)};
\addplot[only marks, mark=*, mark size=2pt, color=cPL, mark options={draw=white, line width=0.4pt}] coordinates {\TCh};
\addlegendentry{H100}
\addplot[only marks, mark=triangle*, mark size=2.5pt, color=cPL!55!white, mark options={draw=cPL, line width=0.5pt}] coordinates {\TCt};
\addlegendentry{RTX 3090}
\node[font=\scriptsize, text=cInk, anchor=north] at (axis cs:0.012,1.4) {proposals};
\node[font=\scriptsize, text=cInk, anchor=east] at (axis cs:0.2,1.62) {row packing};
\end{axis}
\end{tikzpicture}
\caption{\method{}'s lead over the best existing schedule at the 30 Ouro-1.4B points against $\rho$, the share of a pass spent reading one loop's rows, from the engine's calibration. In-row proposals carry the gain at light load, row packing at heavy load (Spearman correlation between $\rho$ and the lead: 0.49).}
\label{fig:rho}
\end{figure}

\begin{figure*}[h]
\centering
\input{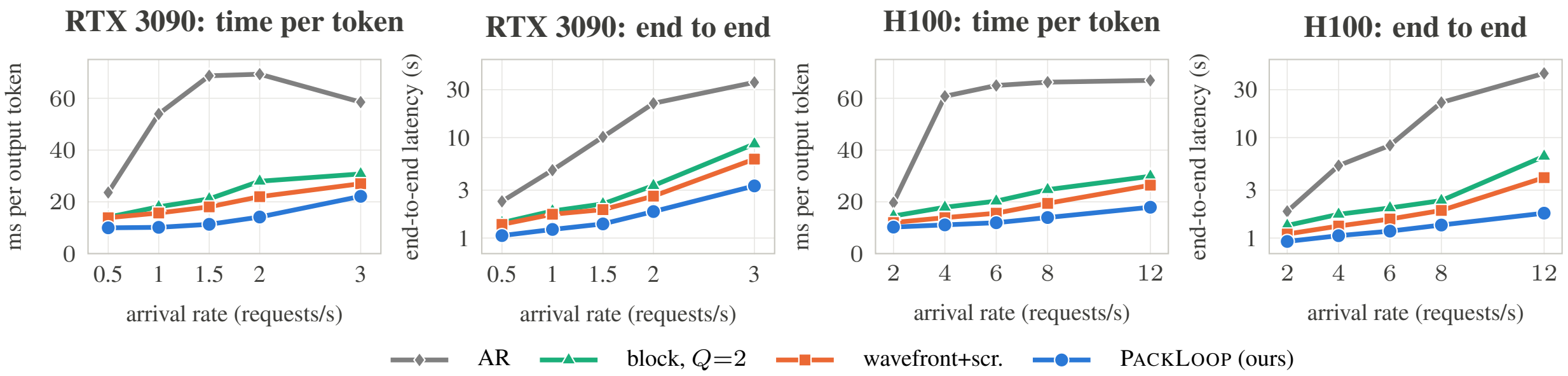}
\begin{tikzpicture}
\pgfplotsset{
  olfig/.style={
    height=3.6cm, width=0.27\textwidth,
    tick label style={font=\scriptsize, color=cInk},
    label style={font=\scriptsize, color=cInk},
    title style={font=\small\bfseries, color=cInk, yshift=-3pt},
    grid=major, grid style={color=cDone!45, line width=0.3pt},
    axis line style={color=cDone}, major tick length=0pt, minor tick length=0pt, xtick pos=bottom, ytick pos=left, clip=false,
    xlabel={arrival rate (requests/s)},
    every axis plot/.append style={line width=1.4pt},
  },
  tpot/.style={ymin=0, ymax=75, ytick={0,20,40,60}, ylabel={ms per output token}},
  e2e/.style={ymin=0.7, ymax=60, ytick={1,3,10,30}, yticklabels={1,3,10,30}, ylabel={end-to-end latency (s)}},
  olar/.style={color=gray, mark=diamond*, mark size=2.2pt, mark options={draw=white, line width=0.4pt}},
  olbq/.style={color=cBQ, mark=triangle*, mark size=2.4pt, mark options={draw=white, line width=0.4pt}},
  olws/.style={color=cWS, mark=square*, mark size=1.8pt, mark options={draw=white, line width=0.4pt}},
  olpl/.style={color=cPL, mark=*, mark size=2.1pt, mark options={draw=white, line width=0.4pt}},
}
\begin{groupplot}[group style={group size=4 by 1, horizontal sep=1.05cm}, olfig]
\nextgroupplot[tpot, title={RTX 3090: time per token}, xmin=0.3, xmax=3.2, xtick={0.5,1,1.5,2,3}, xticklabels={0.5,1,1.5,2,3}]
\addplot[olar] coordinates {\OLtTar}; \addplot[olbq] coordinates {\OLtTbq}; \addplot[olws] coordinates {\OLtTws}; \addplot[olpl] coordinates {\OLtTpl};
\nextgroupplot[ymode=log, e2e, title={RTX 3090: end to end}, xmin=0.3, xmax=3.2, xtick={0.5,1,1.5,2,3}, xticklabels={0.5,1,1.5,2,3}]
\addplot[olar] coordinates {\OLeTar}; \addplot[olbq] coordinates {\OLeTbq}; \addplot[olws] coordinates {\OLeTws}; \addplot[olpl] coordinates {\OLeTpl};
\nextgroupplot[tpot, title={H100: time per token}, xmin=1.3, xmax=12.7, xtick={2,4,6,8,12}]
\addplot[olar] coordinates {\OLtHar}; \addplot[olbq] coordinates {\OLtHbq}; \addplot[olws] coordinates {\OLtHws}; \addplot[olpl] coordinates {\OLtHpl};
\nextgroupplot[ymode=log, e2e, title={H100: end to end}, xmin=1.3, xmax=12.7, xtick={2,4,6,8,12},
  legend style={font=\scriptsize, draw=none, fill=none, at={(-1.45,-0.42)}, anchor=north, legend columns=4, column sep=6pt}, legend cell align=left]
\addplot[olar] coordinates {\OLeHar}; \addlegendentry{AR}
\addplot[olbq] coordinates {\OLeHbq}; \addlegendentry{block, $Q{=}2$}
\addplot[olws] coordinates {\OLeHws}; \addlegendentry{wavefront+scr.}
\addplot[olpl] coordinates {\OLeHpl}; \addlegendentry{\method{} (ours)}
\end{groupplot}
\end{tikzpicture}
\caption{Latency of Ouro-1.4B under open-loop Poisson arrivals: median time per output token and mean end-to-end latency (log scale) against the arrival rate. Mixed GSM8K, MBPP, and HumanEval prompts decoded to their natural end; all schedules in the same engine.}
\label{fig:openloop}
\end{figure*}

\textbf{Open-loop latency.} \Cref{fig:openloop} shows every arrival rate of the open-loop experiment (60\,s of arrivals per rate; 8 request slots on the RTX 3090 with up to 3 requests/s, 32 slots on the H100 with up to 12 requests/s). At most two answers per task and rate differ from AR's, all at BF16 near-ties. AR's time per token saturates at 54--69\,ms, after which requests queue. \method{} has the lowest time per token (10--22\,ms) and the lowest end-to-end latency at every rate, and its end-to-end margin over the screened wavefront grows from 1.2--1.3$\times$ at the lowest rate to 1.8--2.3$\times$ at the highest.

\begin{figure*}[h]
\centering
\input{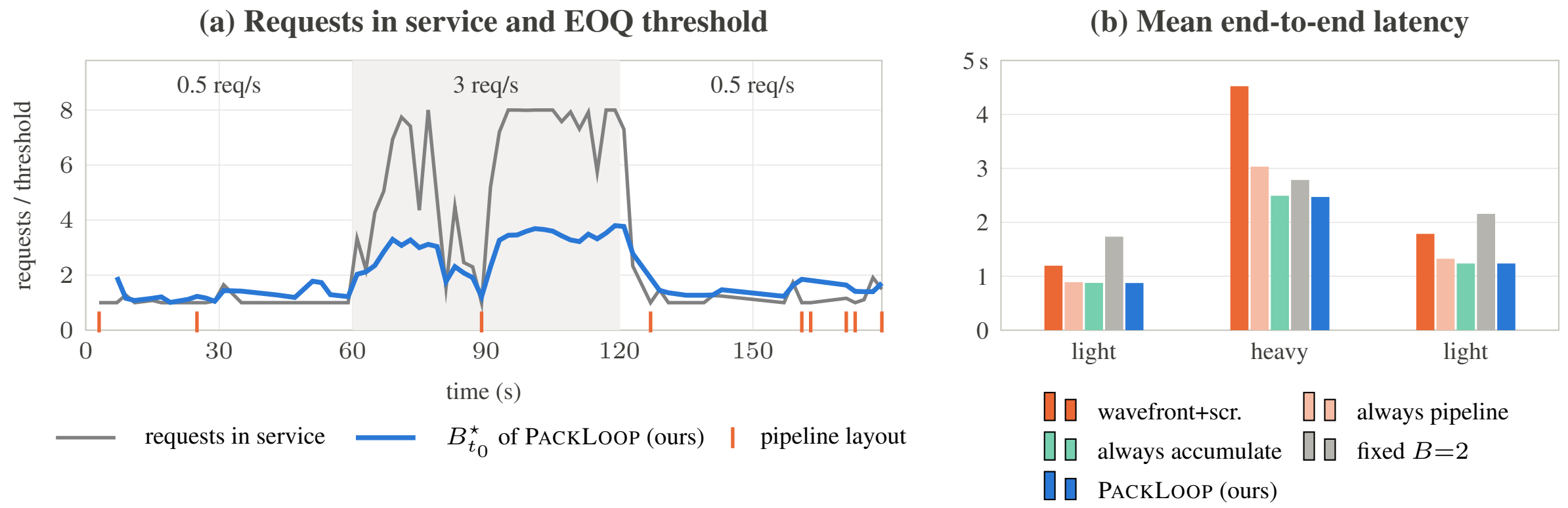}
\begin{tikzpicture}
\pgfplotsset{
  stfig/.style={
    height=4.3cm,
    tick label style={font=\scriptsize, color=cInk}, label style={font=\scriptsize, color=cInk},
    title style={font=\small\bfseries, color=cInk, yshift=-3pt},
    grid=major, grid style={color=cDone!45, line width=0.3pt}, axis line style={color=cDone},
    major tick length=0pt, xtick pos=bottom, ytick pos=left, clip=false,
  },
}
\begin{axis}[stfig, name=sa, width=0.56\textwidth,
  title={(a) Requests in service and EOQ threshold},
  xmin=0, xmax=\ST, ymin=0, ymax=9.8, xtick={0,30,60,90,120,150,180}, ytick={0,2,4,6,8}, ylabel={requests / threshold},
  xlabel={time (s)},
  legend style={font=\scriptsize, draw=none, fill=none, at={(0.5,-0.3)}, anchor=north, legend columns=3, column sep=6pt}, legend cell align=left,
]
\fill[cDone!25] (axis cs:60,0) rectangle (axis cs:120,9.8);
\node[font=\scriptsize, text=cInk, anchor=north] at (axis cs:90,9.7) {3 req/s};
\node[font=\scriptsize, text=cInk, anchor=north] at (axis cs:30,9.7) {0.5 req/s};
\node[font=\scriptsize, text=cInk, anchor=north] at (axis cs:150,9.7) {0.5 req/s};
\addplot[color=gray, line width=1.0pt, mark=none] coordinates {\Sact};
\addlegendentry{requests in service}
\addplot[color=cPL, line width=1.4pt, mark=none] coordinates {\Sthr};
\addlegendentry{$B^\star_{t_0}$ of \method{} (ours)}
\addplot[only marks, mark=|, mark size=3pt, color=cWS, line width=1pt] coordinates {\Snar};
\addlegendentry{pipeline layout}
\end{axis}
\begin{axis}[stfig, name=sb, at={($(sa.east)+(1.2cm,0)$)}, anchor=west, width=0.42\textwidth,
  title={(b) Mean end-to-end latency},
  ybar=0.6pt, bar width=5.2pt, ymin=0, ymax=5.0, ytick={0,1,2,3,4,5}, yticklabels={0,1,2,3,4,5\,s},
  xmin=0.5, xmax=3.5, xtick={1,2,3}, xticklabels={light,heavy,light}, xmajorgrids=false,
  legend style={font=\scriptsize, draw=none, fill=none, at={(0.5,-0.2)}, anchor=north, legend columns=2, column sep=4pt}, legend cell align=left,
]
\addplot[fill=cWS, draw=none] coordinates {\SbarA}; \addlegendentry{\SlabA}
\addplot[fill=cWS!45, draw=none] coordinates {\SbarB}; \addlegendentry{\SlabB}
\addplot[fill=cBQ!60, draw=none] coordinates {\SbarC}; \addlegendentry{\SlabC}
\addplot[fill=cDone!90!black, draw=none] coordinates {\SbarD}; \addlegendentry{\SlabD}
\ifdefined\SbarE
\addplot[fill=cPL, draw=none] coordinates {\SbarE}; \addlegendentry{\SlabE}
\fi
\end{axis}
\end{tikzpicture}
\caption{Open-loop arrivals that step from 0.5 to 3 requests/s and back (Ouro-1.4B, RTX 3090, 8 slots, mixed GSM8K, MBPP, and HumanEval prompts). (a) Requests in service and \method{}'s EOQ threshold $B^\star_{t_0}$ (2\,s bins; orange ticks: bins where the rule pipelines). (b) Mean end-to-end latency of the requests arriving in each phase; AR takes 1.7, 33.7, and 44.6\,s.}
\label{fig:steps}
\end{figure*}

\textbf{Admission under a load step.} \Cref{fig:steps} runs the open-loop server on the RTX 3090 with an arrival rate that steps from 0.5 to 3 requests/s and back, 60\,s per phase, and records the admission state of every pass. With one request in service, $B^\star_{t_0}$ stays near 1.2, so the first deep loop advances almost every pass. When the eight slots fill, the threshold rises to 3.1--3.8 within a few seconds, and it falls back when the load drops. One configuration thus matches the better static choice in every phase: \method{} has the lowest mean end-to-end latency in all three (0.87, 2.47, and 1.24\,s), a fixed cadence $B=2$ is 1.7--2.0$\times$ slower in the light phases and 13\% slower in the heavy one, and always pipelining is 23\% slower under heavy load.

\begin{table*}[h]
\caption{Decode throughput (tokens/s) of Ouro-1.4B. All schedules run in the same engine and job; Ours is \method{}, and ``vs best'' divides it by the per-point fastest existing schedule. vllm-rlt is its fastest mode at that point (H100: measured at $N\in\{1,4,16,64\}$). Bold: \method{} leads by more than 5\%; underlined ratio: within 5\%.}
\label{tab:main}
\centering\small
\setlength{\tabcolsep}{3pt}
\resizebox{\textwidth}{!}{%
\begin{tabular}{ll|rrrrrrrr|rrrrrrrr}
\toprule
& & \multicolumn{8}{c|}{RTX 3090} & \multicolumn{8}{c}{H100 NVL} \\
$P$ & $N$ & AR & Wave & Wave+scr & Blk Q2 & Blk Q4 & Ours & vs best & vs vllm-rlt & AR & Wave & Wave+scr & Blk Q2 & Blk Q4 & Ours & vs best & vs vllm-rlt \\
\midrule
256 & 1 & 59.9 & 142.2 & 158.3 & 129.6 & 121.9 & \textbf{218.8} & 1.38 & 2.21 & 110.4 & 254.5 & 283.5 & 203.5 & 195.5 & \textbf{323.4} & 1.14 & 2.54 \\
256 & 2 & 112.0 & 276.9 & 307.4 & 243.6 & 234.0 & \textbf{418.2} & 1.36 & 2.31 & 215.7 & 531.3 & 586.7 & 402.7 & 393.9 & \textbf{747.6} & 1.27 & -- \\
256 & 4 & 213.4 & 477.1 & 530.6 & 414.5 & 392.5 & \textbf{660.5} & 1.24 & 2.15 & 421.9 & 994.3 & 1104.5 & 763.6 & 705.3 & \textbf{1376.6} & 1.25 & 3.42 \\
256 & 8 & 397.4 & 746.1 & 821.3 & 652.5 & 608.4 & \textbf{951.6} & 1.16 & 2.02 & 813.7 & 1694.7 & 1888.4 & 1328.5 & 1226.2 & \textbf{2168.3} & 1.15 & -- \\
256 & 16 & & & & & & & & & 1477.7 & 2739.7 & 3018.8 & 2181.4 & 1987.6 & \textbf{3403.2} & 1.13 & 3.21 \\
256 & 32 & & & & & & & & & 2510.4 & 3803.3 & 4162.2 & 3265.4 & 2903.1 & \textbf{4581.3} & 1.10 & -- \\
256 & 64 & & & & & & & & & 3891.1 & 4505.1 & 4925.6 & 3963.3 & 3508.1 & \textbf{5622.3} & 1.14 & 2.19 \\
\midrule
1024 & 1 & 56.1 & 118.6 & 131.4 & 109.1 & 103.6 & \textbf{162.3} & 1.24 & 1.87 & 106.1 & 208.2 & 247.0 & 176.5 & 165.0 & 246.9 & \underline{1.00} & 2.18 \\
1024 & 2 & 102.0 & 205.4 & 227.4 & 190.0 & 186.7 & \textbf{280.1} & 1.23 & 1.88 & 200.2 & 450.7 & 486.4 & 325.8 & 316.8 & \textbf{517.7} & 1.06 & -- \\
1024 & 4 & 181.3 & 305.2 & 334.2 & 274.8 & 275.7 & \textbf{427.3} & 1.28 & 2.00 & 384.5 & 777.9 & 866.2 & 590.9 & 557.9 & \textbf{979.3} & 1.13 & 2.84 \\
1024 & 8 & 303.2 & 408.0 & 445.0 & 361.6 & 347.1 & \textbf{591.2} & 1.33 & 1.99 & 713.4 & 1170.4 & 1305.3 & 938.2 & 880.4 & \textbf{1479.7} & 1.13 & -- \\
1024 & 16 & & & & & & & & & 1182.8 & 1551.3 & 1682.1 & 1353.6 & 1226.7 & \textbf{2130.3} & 1.27 & 2.91 \\
1024 & 32 & & & & & & & & & 1744.3 & 1866.1 & 2015.8 & 1633.2 & 1449.1 & \textbf{2863.3} & 1.42 & -- \\
1024 & 64 & & & & & & & & & 2209.5 & 1989.1 & 2156.2 & 1818.2 & 1605.7 & \textbf{3294.4} & 1.49 & 2.30 \\
\midrule
4096 & 1 & 45.4 & 85.0 & 90.1 & 77.7 & 80.2 & \textbf{109.6} & 1.22 & 1.74 & 100.9 & 194.5 & 207.9 & 157.3 & 153.9 & \textbf{225.8} & 1.09 & -- \\
4096 & 2 & 79.5 & 118.5 & 127.1 & 121.5 & 135.3 & \textbf{179.8} & 1.33 & 1.88 & 178.5 & 308.1 & 334.6 & 257.9 & 259.0 & 351.6 & \underline{1.05} & -- \\
4096 & 4 & 119.8 & 137.0 & 147.3 & 135.8 & 148.1 & \textbf{237.5} & 1.60 & 1.80 & 303.7 & 450.9 & 490.0 & 392.9 & 401.2 & \textbf{611.9} & 1.25 & -- \\
4096 & 8 & & & & & & & & & 490.2 & 534.7 & 576.3 & 472.3 & 460.6 & \textbf{837.2} & 1.45 & -- \\
4096 & 16 & & & & & & & & & 623.4 & 569.1 & 627.9 & 515.0 & 466.2 & \textbf{968.1} & 1.54 & -- \\
\bottomrule
\end{tabular}}
\end{table*}

\begin{table}[h]
\caption{\ours{} against re-implementations of the concurrent LoopSkim and LoopFill in our engine, in the same jobs as \cref{tab:main,tab:26b}. LoopSkim-style: drafts copied on suffix matches of at least two tokens, checked at every loop, faster of 2 and 4 draft tokens per point, copied positions sharing a row. LoopFill layout: the same drafts with one row per draft position at every loop. Throughput ratios, range (median); sig.: points where the paired 95\% interval lies above 1.}
\label{tab:concurrent}
\centering\small
\setlength{\tabcolsep}{4pt}
\begin{tabular}{llrcrcr}
\toprule
Model & GPU & Pts & vs LoopSkim-style & sig. & vs LoopFill layout & sig. \\
\midrule
Ouro-1.4B & RTX 3090 & 11 & 1.01--1.37 (1.07) & 10 & 1.46--3.27 (1.82) & 11 \\
Ouro-1.4B & H100 & 18 & 0.99--1.37 (1.06) & 11 & 1.21--3.25 (1.74) & 17 \\
Ouro-2.6B & RTX 3090 & 8 & 1.02--1.19 (1.07) & 7 & 1.44--2.34 (1.76) & 8 \\
Ouro-2.6B & H100 & 17 & 1.00--1.39 (1.08) & 15 & 1.37--3.29 (1.82) & 17 \\
\bottomrule
\end{tabular}
\end{table}

\textbf{LoopSkim with its trained draft head.}
\label{app:dhead}
LoopSkim can also draft with a trained head: one causal decoder layer (16.8M parameters) that reads the last 32 loop-1 states with the embedding of each next token and predicts the next 8 loop-1 states in one pass. We reproduce it following \citet{loopskim}: the same mix of 9{,}025 training prompts (GSM8K, MATH, CodeAlpaca, BBH, and UltraChat; none of our evaluation sets), the model's own greedy continuations, a cross-entropy loss on the loop-1 readout plus 0.2 times a smooth-L1 loss on the states, and 16{,}000 steps (five minutes on one RTX 4090). On held-out continuations, the head's first draft token matches the final output for 67\% of tokens with Ouro-1.4B and 74\% with Ouro-2.6B. In the engine the head supplies a proposal whenever the copy lookup finds no match of at least two tokens, LoopSkim's chain selection; its tokens are verified like copied ones, so outputs stay exact (FP32 checks pass). The LoopSkim-style schedule takes the fastest of six head settings at each point (2, 4, or 8 draft tokens; copy first or head only), whereas \method{} uses one setting with head proposals of up to 3 tokens. \Cref{tab:dhead} shows that the head speeds up both schedules, LoopSkim more because \method{} already fills its rows with copied proposals. With the same drafter, the two schedules are at parity at light load, and \method{} leads where reads dominate.

\begin{table}[h]
\caption{\ours{} against LoopSkim with its trained draft head, both schedules drafting with the same head (8 prompt sets, same job). Head gain: LoopSkim-style with the head over without, median. Lead: \method{} with the head over the per-point fastest LoopSkim-style head setting, range (median); sig.: points whose paired 95\% interval lies above / below 1. Read-heavy: points with $NP\ge 8$K.}
\label{tab:dhead}
\centering\small
\setlength{\tabcolsep}{4pt}
\begin{tabular}{llrcccc}
\toprule
Model & GPU & Pts & head gain & lead of \ours{} & sig.\ $+$\,/\,$-$ & lead, read-heavy \\
\midrule
Ouro-1.4B & RTX 3090 & 11 & 1.10 & 0.88--1.18 (1.00) & 2\,/\,1 & 1.05--1.18 \\
Ouro-1.4B & RTX 4090 & 11 & 1.14 & 0.92--1.18 (1.01) & 3\,/\,0 & 1.07--1.18 \\
Ouro-2.6B & RTX 3090 & 8 & 1.20 & 0.86--1.02 (0.97) & 0\,/\,1 & -- \\
\bottomrule
\end{tabular}
\end{table}

\textbf{Exactness in FP32.} \Cref{tab:fp32} lists the FP32 checks (TF32 disabled, Ouro-1.4B, greedy). In every check, every request's output equals the AR output token for token. The early-exit check compares tokens and per-token exit loops with a cell-by-cell implementation of the exit rule that does not use the engine.

\begin{table}[h]
\caption{FP32 exactness checks. Every configuration reproduces the AR output (or, for early exit, the reference implementation) token for token on every request.}
\label{tab:fp32}
\centering\small
\resizebox{\linewidth}{!}{%
\begin{tabular}{p{0.52\linewidth}lll}
\toprule
Configurations & GPU & Context; requests & Tokens/req. \\
\midrule
\method{}; one row per position; CopyLoop layout; proposals verified round by round on AR; block with one row per position & RTX 3090 & 256; 1, 8 & 64 \\
\method{} with $f^\star=0.37$ and $0.40$, including a mid-run switch from pipeline to blocks & RTX 3090 & 256; 1, 4, 8 and 1024; 4 & 128 \\
Fused shared block and merged attention, 3 schedules & RTX 3090 & 256, 1024; 1, 4 & 64 \\
Early exit (CDB rule): AR, blocks $Q{=}2,4$; tokens and exit loops & RTX 3090 & 256; 8 & 128 \\
Sampling at temperature 0.7 and 1.0 against AR with the same noise: \method{}, screened wavefront, blocks $Q{=}4$ & RTX 3090 & 256; 1, 8 & 64 \\
Draft-head proposals (\cref{app:dhead}): \method{} and LoopSkim-style, copy first or head only, Ouro-1.4B and 2.6B & RTX 3090 & 256; 1, 8 & 64 \\
\bottomrule
\end{tabular}}
\end{table}

\textbf{Engine.} The engine keeps a dense KV cache per loop and captures one CUDA graph per pass shape, that is, per combination of widths of the advancing loops. On the RTX 3090, narrow matrix multiplications ($M\le 64$) use a Triton kernel that runs within 1.13--1.20$\times$ of the bandwidth bound of reading the weights once. Every baseline schedule uses the same engine settings as \method{}.

\textbf{Load-set width.} With the load threshold $w_n$ (4 on the RTX 3090, 8 on the H100 and the RTX 4090), the loop-1 width rule is within 2.3\% of the better fixed width at every RTX 3090 and H100 point and within 5.3\% on the RTX 4090.

\textbf{External protocols.} With one request and CopyLoop's own inputs, generation lengths, and engine on the same RTX 3090, CopyLoop runs at 211.3 / 364.5 tokens/s on GSM8K / HumanEval and \method{} at 242.4 / 436.6 (1.15$\times$ / 1.20$\times$), using 0.78 / 0.43 passes per token. On the same protocol, LoopSpec's best mode runs at 165.4 / 190.6 tokens/s, so \method{} is 1.47$\times$ / 2.29$\times$ faster. WaveFront's released benchmark, run at acceptance rate 1 (its upper bound), reaches 106.3 / 66.0 / 26.1 tokens/s at contexts 256 / 1024 / 4096; \method{} is 2.1--4.2$\times$ faster at the same contexts with real acceptance. Part of the WaveFront gap is its engine, whose AR decoding runs at 42\% of ours at context 4096 (18.9 against 45.5 tokens/s). Under CDB's exit rule (threshold 0.2, signal one loop late) and identical inputs, \method{} is 1.89--2.60$\times$ faster than CDB in vllm-rlt on the RTX 3090 and 2.73--3.59$\times$ on the H100. End to end, \method{} is 1.74--2.31$\times$ faster than the fastest vllm-rlt mode on the RTX 3090 (1.99--2.36$\times$ for Ouro-2.6B) and 2.18--3.42$\times$ on the H100.

\textbf{Engine components.} Turning off engine components one at a time (fusion, packing, merged attention, and narrow GEMMs on the RTX 3090; context 1024; 1, 8, and 64 requests) keeps \method{} 1.16--1.39$\times$ faster than the screened wavefront on the RTX 3090; on the H100, without fusion or merged attention, it stays 1.00--1.51$\times$ faster. Packing matters most: removing it costs \method{} 20--22\% on the RTX 3090 and 32--34\% at 1 and 8 requests on the H100, because without it the linear layers also compute the padding of \method{}'s wider rows.

\textbf{Row layout on other GPUs.} The layout ablation of \cref{tab:reads} also holds on bandwidth-poorer GPUs. With an earlier version of the engine (blocks of $Q=4$, no proposals), splitting each block row into one row per position lowers the speedup over AR as follows: on an A6000 at context 1024 from 1.83$\times$ to 1.07$\times$ with 16 requests and from 1.66$\times$ to 0.91$\times$ with 32; on an A100 at context 1024 from 1.97$\times$ to 1.22$\times$ (16 requests), 1.73$\times$ to 1.00$\times$ (32), and 1.60$\times$ to 0.88$\times$ (64), and at context 4096 with 16 requests from 1.87$\times$ to 0.95$\times$. With one row per position, larger blocks are worse (on the A6000 at context 1024, $Q=8$ with one row per position runs at 0.76--0.98$\times$ of AR), which is why per-token-row systems shrink the draft length under load \citep{vllmrlt}; with shared rows, $Q=4$--$8$ is best.

\textbf{Bandwidth.} Fitting the RTX 3090 row/cell microbenchmark of \cref{fig:teaser}(a) gives 0.26\,$\mu$s per row and context position. A row of one loop holds 192\,KiB of KV per position for Ouro-1.4B (24 layers, width 2048), so reading a row runs at about 760\,GB/s, 81\% of the GPU's 936\,GB/s. Rows are therefore bandwidth-bound in our engine, and a schedule saves time only by reading fewer of them.

\textbf{AR on the H100.} Our AR decoding is 1.9--3.3$\times$ faster than vllm-rlt without speculation at the H100 points of \cref{tab:main}. With one request, a pass over one loop replays 442 kernels in 2.15\,ms at context 256 (2.24\,ms at 1024): the weight GEMMs take 1.1\,ms, attention 0.23\,ms, and the rest is spread over small per-layer kernels, so at this load the H100 is limited by kernel count more than by bandwidth.

\textbf{Ouro-2.6B per point.} \Cref{tab:26b} gives the Ouro-2.6B points (tokens/s; RTX 3090 with 8 prompt sets, H100 with 4; on each GPU all schedules of a row run in the same job). Every existing schedule ran in the same job, and the screened wavefront was the fastest of them at every point on both GPUs; plain wavefront and blocks of $Q=2,4$ were slower.

\begin{table}[h]
\caption{Ouro-2.6B (tokens/s). $\times$best: \method{} over the fastest existing schedule (the screened wavefront at every point).}
\label{tab:26b}
\centering\small
\setlength{\tabcolsep}{4pt}
\begin{tabular}{lrrrrrc}
\toprule
GPU & $P$ & $N$ & AR & Wave+scr & \ours{} & $\times$best \\
\midrule
RTX 3090 & 256 & 1 & 31.7 & 92.9 & 122.2 & 1.32 \\
 & 256 & 2 & 59.2 & 174.2 & 243.5 & 1.40 \\
 & 256 & 4 & 112.5 & 295.9 & 384.6 & 1.30 \\
 & 256 & 8 & 209.2 & 448.8 & 519.5 & 1.16 \\
 & 1024 & 1 & 29.5 & 79.2 & 111.8 & 1.41 \\
 & 1024 & 2 & 53.6 & 135.7 & 185.9 & 1.37 \\
 & 1024 & 4 & 94.9 & 196.9 & 281.1 & 1.43 \\
 & 4096 & 1 & 23.6 & 53.1 & 67.5 & 1.27 \\
\midrule
H100 NVL & 256 & 1 & 60.0 & 176.8 & 217.5 & 1.23 \\
 & 256 & 2 & 114.5 & 344.0 & 422.2 & 1.23 \\
 & 256 & 4 & 225.7 & 641.3 & 814.8 & 1.27 \\
 & 256 & 8 & 435.3 & 1086.9 & 1307.4 & 1.20 \\
 & 256 & 16 & 793.7 & 1684.9 & 1963.9 & 1.17 \\
 & 256 & 32 & 1342.1 & 2263.9 & 2517.0 & 1.11 \\
 & 256 & 64 & 2065.0 & 2666.7 & 3354.4 & 1.26 \\
 & 1024 & 1 & 57.7 & 157.1 & 191.1 & 1.22 \\
 & 1024 & 2 & 106.5 & 303.9 & 364.7 & 1.20 \\
 & 1024 & 4 & 203.7 & 524.3 & 661.8 & 1.26 \\
 & 1024 & 8 & 378.5 & 766.6 & 926.8 & 1.21 \\
 & 1024 & 16 & 626.1 & 968.9 & 1323.9 & 1.37 \\
 & 1024 & 32 & 906.4 & 1132.6 & 1705.0 & 1.51 \\
 & 4096 & 1 & 53.6 & 125.0 & 138.4 & 1.11 \\
 & 4096 & 2 & 94.4 & 202.2 & 226.9 & 1.12 \\
 & 4096 & 4 & 159.5 & 269.9 & 384.8 & 1.43 \\
 & 4096 & 8 & 247.1 & 328.3 & 530.4 & 1.62 \\
\bottomrule
\end{tabular}
\end{table}

\textbf{RTX 4090.} \Cref{tab:4090} gives the Ouro-1.4B grid on an RTX 4090 (8 prompt sets, all schedules of a row in the same job). The engine settings are the H100's (no narrow-GEMM kernel) and are the same for every schedule. The only per-GPU constant, $w_n=8$, comes from the same short sweep as on the other GPUs, on MATH-500 prompts: with fewer than 8 requests the wide loop-1 row is faster at both contexts, with 8 the two widths tie (context 256) or the narrow one is 4\% faster (context 1024), and with 16 the narrow one is 5\% faster. The screened wavefront is the fastest existing schedule at short context, and blocks of $Q=4$ take over at context 4096 with 2--4 requests; \method{} leads both, by most where reads dominate: at contexts 1024 and 4096 the lead grows with concurrency, up to 1.68$\times$ with 4 requests at 4096.

\begin{table}[h]
\caption{Ouro-1.4B on the RTX 4090 (tokens/s). $\times$best: \method{} over the fastest existing schedule; every lead is significant (paired 95\% interval, 8 sets).}
\label{tab:4090}
\centering\small
\setlength{\tabcolsep}{3.5pt}
\begin{tabular}{rrrrrrrrc}
\toprule
$P$ & $N$ & AR & Wave & Wave+scr & Blk $Q{=}2$ & Blk $Q{=}4$ & \ours{} & $\times$best \\
\midrule
256 & 1 & 67.3 & 154.0 & 169.5 & 138.4 & 138.0 & \textbf{236.5} & 1.40 \\
256 & 2 & 117.0 & 293.1 & 325.5 & 266.1 & 254.0 & \textbf{473.0} & 1.45 \\
256 & 4 & 223.3 & 517.0 & 569.0 & 468.8 & 446.0 & \textbf{774.6} & 1.36 \\
256 & 8 & 415.2 & 840.3 & 930.0 & 759.7 & 724.2 & \textbf{1205.6} & 1.30 \\
1024 & 1 & 63.7 & 131.9 & 139.8 & 116.9 & 115.7 & \textbf{174.7} & 1.25 \\
1024 & 2 & 107.9 & 223.9 & 247.3 & 212.4 & 208.0 & \textbf{333.8} & 1.35 \\
1024 & 4 & 194.4 & 335.4 & 370.8 & 316.5 & 317.3 & \textbf{529.0} & 1.43 \\
1024 & 8 & 328.8 & 469.2 & 508.6 & 430.0 & 423.2 & \textbf{779.3} & 1.53 \\
4096 & 1 & 55.3 & 93.2 & 99.0 & 91.6 & 93.5 & \textbf{129.3} & 1.31 \\
4096 & 2 & 83.0 & 127.7 & 139.9 & 138.7 & 149.2 & \textbf{217.0} & 1.45 \\
4096 & 4 & 129.6 & 153.5 & 165.5 & 153.0 & 172.2 & \textbf{290.0} & 1.68 \\
\bottomrule
\end{tabular}
\end{table}

\end{document}